\documentclass[journal]{IEEEtran}

\ifCLASSOPTIONcompsoc
  \usepackage[nocompress]{cite}
\else
  \usepackage{cite}
\fi

\usepackage[usenames,dvipsnames]{color}

\usepackage{pgfplots}
\pgfplotsset{compat=newest}

\usepackage{enumitem}
\usepackage{epsfig}
\usepackage{graphicx}
\usepackage{amsmath}
\usepackage{amsfonts}
\usepackage{amssymb}
\usepackage{array}
\usepackage{multirow}
\usepackage{rotating}
\usepackage{epstopdf}
\usepackage{caption}
\usepackage{subcaption}
\usepackage{xcolor}
\usepackage{adjustbox}

\usepackage[makeroom]{cancel}

\usepackage{url}
\usepackage{afterpage}
\usepackage{booktabs}

\newcolumntype{Y}{>{\centering\arraybackslash} m{0.8cm} }
\newcolumntype{R}{>{\centering\arraybackslash} m{0.7cm} }

\begin{document}

\title{Biometric Presentation Attack Detection:\\Beyond the Visible Spectrum}

\author{Ruben~Tolosana, Marta~Gomez-Barrero, Christoph~Busch, Javier~Ortega-Garcia~\IEEEmembership{Life~Fellow,~IEEE}
\thanks{R. Tolosana and J. Ortega-Garcia are with the Biometrics and Data Pattern Analytics (BiDA) Lab, Universidad Autonoma de Madrid, Spain (e-mail: \{ruben.tolosana,javier.ortega\}@uam.es).}
\thanks{M. Gomez-Barrero and C. Busch are with the da/sec - Biometrics and Internet Security Research Group, Hochschule Darmstadt, Germany (e-mail: \{marta.gomez-barrero,christoph.busch\}@h-da.de).}}%

%
%

\markboth{}
{}
%



\maketitle

\begin{abstract}
The increased need for unattended authentication in multiple scenarios has motivated a wide deployment of biometric systems in the last few years. This has in turn led to the disclosure of security concerns specifically related to biometric systems. Among them, Presentation Attacks (PAs, i.e., attempts to log into the system with a fake biometric characteristic or presentation attack instrument) pose a severe threat to the security of the system: any person could eventually fabricate or order a gummy finger or face mask to impersonate someone else. The biometrics community has thus made a considerable effort to the development of automatic Presentation Attack Detection (PAD) mechanisms, for instance through the international LivDet competitions. 

In this context, we present a novel fingerprint PAD scheme based on $i)$ a new capture device able to acquire images within the short wave infrared (SWIR) spectrum, and $ii)$ an in-depth analysis of several state-of-the-art techniques based on both handcrafted and deep learning features. The approach is evaluated on a database comprising over 4700 samples, stemming from 562 different subjects and 35 different presentation attack instrument (PAI) species. The results show the soundness of the proposed approach with a detection equal error rate (D-EER) as low as 1.36\% even in a realistic scenario where five different PAI species are considered only for testing purposes (i.e., unknown attacks).


\end{abstract}

\begin{IEEEkeywords}
Presentation Attack Detection, Biometrics, Deep Learning, CNN, SWIR, Fingerprint
\end{IEEEkeywords}

%
\IEEEpeerreviewmaketitle

\setlength{\tabcolsep}{4.0pt}

\section{Introduction}
\label{sec:intro}



There is an increasing demand in the current society for automatic and reliable authentication of individuals in a wide number of scenarios. To address this need, biometric recognition systems based on the individuals' biological (e.g., iris or fingerprint) or behavioural (e.g., signature or voice) characteristics have been consolidated as a reliable paradigm in the last decades. Its advantages over traditional authentication methods (e.g., no need to carry tokens or remember passwords, they are harder to circumvent and provide at the same time a stronger link between the subject and the action or event), have allowed a wide deployment of biometric systems, including large-scale national and international initiatives such as the Unique ID program of the Indian government \cite{indianUID} or the Smart Border project of the European Comission \cite{SmartBorders}.

In spite of their numerous advantages, biometric systems are vulnerable to external attacks as any other security-related technology. Among all possible attack points defined in \cite{Zwiesele-BioIS-Study-IEEE-CCST2000,Ratha-EnhancingSecurityAndPrivacy-IBM2001,ISO-IEC-30107-1-PAD-Framework-160115}, the biometric capture device is probably the most exposed one: an eventual attacker requires no knowledge about the inner functioning of the system in order to launch an attack and break the system. Instead, he/she can simply present the capture device with a \textit{presentation attack instrument} (PAI), such as a gummy finger or a fingerprint overlay, in order to interfere with its intended behaviour. The main goal might be to impersonate someone else (i.e., active impostor) or to avoid being recognised (i.e., identity concealer). These attacks are known in the ISO/IEC 30107 \cite{ISO-IEC-30107-1-PAD-Framework-160115} as \textit{presentation attacks} (PAs).

Given the severe security threat posed by such PAs, the development of automatic techniques which are able to distinguish between bona fide (i.e., real or live) presentations and access attempts carried out by means of PAIs has become of the utmost importance \cite{Marcel-handbookAntispoofing-2014,hadid15SPMspoofing}. Referred to as \emph{presentation attack detection} (PAD) methods, research in this area has been recently funded by several international projects like the European Tabula Rasa \cite{TabulaRasa} and BEAT \cite{BEAT}, or the more recent US ODIN research program \cite{odinThorProgram}. Together with the organisation of the LivDet -- liveness detection competition series on iris and fingerprint \cite{Ghiani-ReviewFpLivDet-IVC-2017,Mura-LivDet2017-ICB-2018}, where the number of participants has been increasing year after year (up to 17 algorithms submitted in 2017), these initiatives have fostered a considerable number of publications on PAD for different biometric characteristics, including iris \cite{galbally-PADIrisChapter-2017}, fingerprint \cite{Marasco-PAD-SurveyFingerprint-CSUR-2015,Sousedik-PAD-Survey-IET-BMT-2014}, face \cite{galbally-PADfaceSurvey-Access-2014}, or handwritten signature \cite{Tolosana-SignPAD-HoBAS2-2018}.

The initial approaches to PAD were based on the so-called handcrafted features, such as texture descriptors or motion analysis \cite{Marcel-handbookAntispoofing-2014,Raghu-PAD-VeinMotionMagnification-BTAS-2015}. However, in the last years deep learning (DL) has become a thriving topic \cite{Goodfellow-et-al-2016,Deep_NLP,Zhou_2016_CVPR}, and biometric recognition in general, and PAD in particular, are not an exception. DL allows expert systems to learn from experience and understand the world in terms of a hierarchy of simpler units, thereby enabling significant advances in complex domains. The main reasons to understand its high deployment lie on the increasing amount of available data and the evolution of graphical processing units (GPU), which in turn allows the successful training of deep architectures. However, the belief that DL schmes can be only used for tasks with massive amounts of available data is changing thanks to the development of pre-trained models. This transfer learning concept refers to network models that are trained for a given task with large available databases, including any kind of images and not only those expected for the problem at hand. Those models are subsequently retrained (a.k.a. fine-tuned, adapted) for a different task for which data are usually scarce. 

All the aforementioned advances have allowed the deployment of DL architectures in many different fields, including biometric recognition \cite{Rattani-CNNocular-IJCB-2017,2018_IEEEAccess_RNN_Tolosana}. More specifically, convolutional neural networks (CNNs) and deep belief networks (DBNs) have been used for fingerprint PAD purposes, based either on the complete fingerprint samples \cite{Nogueira-PADfingerprintCNN-TIFS-2016,Jang-fingerprintPADcontrastCNN-ICISA-2017,Kim-fingerprintPAD-DNN-PRL-2016} or on a patch-wise manner \cite{Toosi-fingerprintPADpatchCNN-ICCI-2017,Souza-FingerprintPAD-DBM-IJCNN-2017,chugh-fingerprintPADcnn-TIFS-2018}. 

\begin{figure*}[t]
\centering
 \centerline{\includegraphics[width=.99\linewidth]{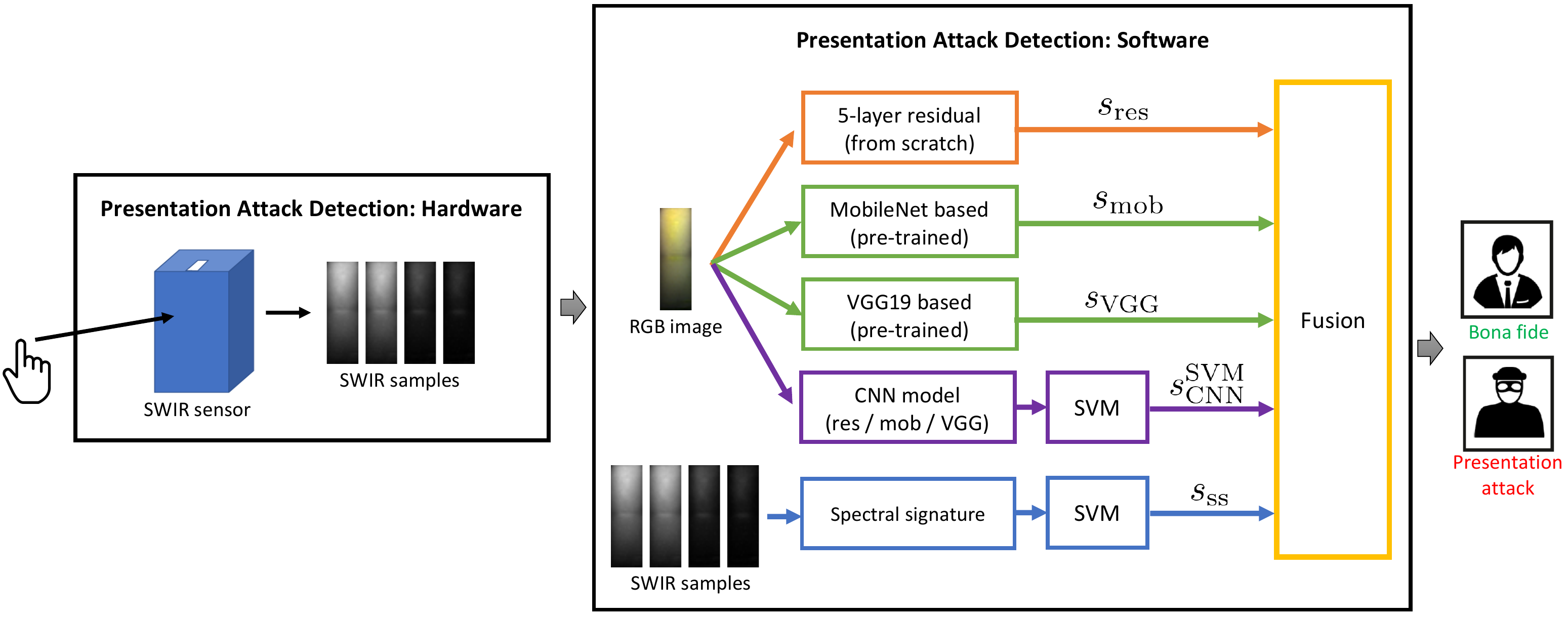}}
 \caption{General diagram of the PAD method proposed. On the left hand, the capture device acquires the samples at four different wavelengths within the SWIR spectrum. On the right hand, several software approaches have been proposed, namely: $i)$ three different state-of-the-art CNN architectures have been tested as an end-to-end solution, $ii)$ the features output by the CNNs have been used to feed an SVM, $iii)$ handcrafted features (i.e., spectral signatures) have been extracted, and $iv)$ a final fusion of the aforementioned algorithms has been evaluated for completeness.} \label{fig:diagram}
\end{figure*}

As it will be described in more detail in Sect.~\ref{sec:related}, DL based PAD approaches have boosted the performance over common PAD benchmarks from the LivDet competitions, achieving detection rates over 90\%. Such high accuracy rates indicate the valuable contributions of the existing approaches. However, the LivDet databases comprise altogether up to 11 different materials for the fabrication of PAIs, even if the choice for the attacker is much wider based on commercial products readily available even online. As a consequence, other databases, comprising a larger number of materials for the fabrication of the PAIs, should be explored. Very few works have considered this issue, including a database comprising over twelve different PAI species in \cite{chugh-fingerprintPADcnn-TIFS-2018}, and 21 materials in \cite{Kanich-fingerprintPAIs-IWBF-2018}. We address this issue with the acquisition of a database including 35 different PAI species, within the US ODIN research program \cite{odinThorProgram}.

In addition, one question that remains mostly unanswered is the following one: Once a artificial neural network is trained on a large number of PAI species, will unknown attacks also be detected? The evaluation carried out in \cite{Kanich-fingerprintPAIs-IWBF-2018} has shown that the error rates were multiplied by a factor of six when unknown PAI species are tested, with respect to the detection accuracy reached on known attacks. Therefore, we can conclude that additional research efforts are needed in this area. To further tackle these issues and in order to reach robustness to unknown attacks, some researchers have considered other sources of information different from traditional capture devices \cite{galbally-PADIrisChapter-2017,Sousedik-PAD-Survey-IET-BMT-2014}. More specifically, the use of multi-spectral near infrared (NIR) technologies has been studied for face \cite{Wang-multispecFacePAD-ACPR-2013,Steiner-facePADswir-Sensors-2016} and fingerprint \cite{Rowe-multispec-fingerprint-book-2008,Chang-FpPANIR-InTech-2011}. 

In this new context, a recent trend for both biometric PAD and face recognition enhancement is based on skin detection. On the one hand, non-skin materials (e.g., a mask or a scarf) can be masked for recognition purposes. On the other hand, such materials can be considered an attempt of a PA. This will be the fundamental idea followed in this article: PAD is regarded as the problem of discriminating skin vs. non-skin materials. In order to overcome one of the main challenges of skin detection, namely, the plurality of different skin colours \cite{Lumini-SkinDetectionComparison-arxiv-2018}, we choose the short wave infrared (SWIR) band as a promising information source. It has been shown that human skin shows characteristic remission properties for multi-spectral SWIR wavelengths, which are independent of a capture subject's age, gender or skin type \cite{Jacquez-spectralSkin-JAP-1955}. In fact, several approaches have been proposed for face recognition in the infrared domain \cite{Ghias-IRfaceRecognition-Survey-PR-2014,Bourlai-FaceRecognitionImageSpectrum-Book-2016}. In particular, for surveillance purposes, the SWIR range has been analysed by several research groups, either as solely source of information or in combination with visible light images \cite{Bourlai-faceRecSWIR-ICPR-2010,Nicolo-SWIR-VIS-FaceRec-TIFS-2012,Narang-SWIR-faceRecognition-IVC-2015}. The advantages of SWIR are mostly its robustness in challenging environmental conditions (e.g., with fog or at night time). In addition, the benefits of multi-spectral hand based recognition within the SWIR bands were studied in \cite{Ferrer-SWIRhand-IS-2014}, where the authors outperformed state-of-the-art recognition approaches. 

For the particular task of PAD, the characteristic remission properties of the human skin observed in the multi-spectral SWIR band were exploited in \cite{Steiner-facePADswir-Sensors-2016} for facial PAD, achieving a 99\% detection accuracy. A similar approach was analysed in \cite{Gomez-Barrero-SWIR-SS-PAD-NISK-2018} over a small fingerprint database, comprising 60 samples. It was shown that the method was able to detect all 12 PAIs except for one. In addition, a preliminary DL approach based on a pre-trained CNN was tested on the same database in \cite{Tolosana-FingerPAD-CNN-SWIR-BIOSIG-2018}, achieving perfect detection rates over the small preliminary database.

Keeping these thoughts in mind, we propose in this work a biometric presentation attack detection method based on SWIR images and state-of-the-art CNN architectures, as depicted in Fig.~\ref{fig:diagram}. Both networks trained from scratch (i.e., a residual network \cite{DBLP:journals/corr/HeZRS15}) and also pre-trained models (i.e., MobileNet \cite{DBLP:journals/corr/HowardZCKWWAA17} and VGG19 \cite{VGG19_2015}) have been analysed. In addition, two different approaches have been studied: $i)$ using the CNNs as an end-to-end solution, and $ii)$ utilising the CNNs as a feature extractor and carrying out classification with support vector machines (SVMs). The results obtained are compared to the handcrafted feature extraction approach proposed in \cite{Gomez-Barrero-SWIR-SS-PAD-NISK-2018}. Then, a final fusion of the different single algorithms is also explored for completeness. The experimental evaluation is carried out on a database captured within the BATL project of the ODIN Program, which includes more than 4700 samples and 35 different PAI species. Over this database, under the unkown attack scenario, a Detection Equal Error Rate (D-EER) of 1.36\% has been achieved, thereby proving the soundness of the proposed approach.

It should be finally noted that, being a skin detection based method, the proposed PAD technique can be applied not only to fingerprints but also to other biometric characteristics, such as the face, the hand, or the periocular regions.

The main contributions of this article can be summarised as follows:
\begin{itemize}
\item Review of the state-of-the-art on fingerprint PAD based on either $i)$ non-conventional capture devices, or $ii)$ traditional sensors and deep learning approaches.

\item Evaluation of multiple state-of-the-art CNN architectures, using both pre-trained models and training the networks from scratch. The CNNs are evaluated as wither end-to-end solutions or alternatively as feature extractors in combination with SVMs.


\item Benchmark of deep learning approaches with high-performing handcrafted features \cite{Gomez-Barrero-SWIR-SS-PAD-NISK-2018}.

\item Fusion of handcrafted and deep learning features on SWIR images.

\item Detection performance evaluation on a large database comprising 35 different PAIs and over 4700 samples.

\item Detection performance evaluation including unknown attacks, achieving a state-of-the-art detection performance.
\end{itemize}

The rest of the article is organised as follows. Sect.~\ref{sec:defs} presents the main terms which will be used in the remainder of the article. Related works on fingerprint PAD are summarised in Sect.~\ref{sec:related}. Sects.~\ref{sec:swir}~and~Sect.~\ref{sec:pad} describe the proposed approach. The evaluation framework is presented in Sect.~\ref{sec:setup}, and the results discussed in Sect.~\ref{sec:res}. Final conclusions are drawn in Sect.~\ref{sec:conc}.







\section{Definitions}
\label{sec:defs}


In the following, we include the main definitions stated within the ISO/IEC 30107-3 standard on biometric presentation attack detection - part 3: testing and reporting \cite{ISO-IEC-30107-3-PAD-metrics-170227}, which will be used throughout the article:

\textbf{Bona fide presentation}: \lq\lq \emph{interaction of the biometric capture subject and the biometric data capture subsystem in the fashion intended by the policy of the biometric system}''. That is, a normal or genuine presentation. 

\textbf{Presentation attack (PA)}: \lq\lq \emph{presentation to the biometric data capture subsystem with the goal of interfering with the operation of the biometric system}''. That is, an attack carried out on the capture device to either conceal your identity or impersonate someone else. 

\textbf{Presentation attack instrument (PAI)}: \lq\lq \emph{biometric characteristic or object used in a presentation attack}''. For instance, a silicone 3D mask or an ecoflex fingerprint overlay.

\textbf{PAI species}: \lq\lq \emph{class of presentation attack instruments created using a common production method and based on different biometric characteristics}''. 

\vspace*{0.2cm}
In order to evaluate the vulnerabilities of biometric systems to PAs, the following metrics should be used:

\textbf{Attack Presentation Classification Error Rate (APCER)}: \lq\lq \emph{proportion of attack presentations using the same PAI species incorrectly classified as bona fide presentations in a specific scenario}''.

\textbf{Bona fide Presentation Classification Error Rate (BPCER)}: \lq\lq \emph{proportion of bona fide presentations incorrectly classified as presentation attacks in a specific scenario}''.

Derived from the aforementioned metrics, the detection equal error rate (D-ERR) is defined as the error rate at the operating point where APCER = BPCER.

\section{Related Works}
\label{sec:related}

In this section, we summarise the key works on fingerprint PAD for both non-conventional optical or capacitive sensors (see Sect.~\ref{sec:related:nc} and Table~\ref{tab:sotaOther}) and using DL approaches on conventional sensors (see Sect.~\ref{sec:related:dl} and Table~\ref{tab:sotaDL}). For further details on fingerprint PAD, the reader is referred to \cite{Sousedik-PAD-Survey-IET-BMT-2014,Marasco-PAD-SurveyFingerprint-CSUR-2015}.

It should be noted that, in addition to the metrics defined in Sect.~\ref{sec:defs} two different metrics are used in the LivDet competitions \cite{Ghiani-ReviewFpLivDet-IVC-2017,Mura-LivDet2017-ICB-2018}. The Average Classification Error Rate (ACER) is defined as the average of the APCER and the BPCER for a pre-defined decision threshold $\delta$:
\begin{equation}
\mathrm{ACER}\left(\delta\right) = \frac{\mathrm{APCER}\left(\delta\right)  + \mathrm{BPCER}\left(\delta\right)}{2}
\end{equation}

It should be noted that averaging APCER and BPCER has been deprecated in ISO/IEC 30107-3. The ACER is reported here for the only purpose to relate our results to the LiveDet competition, where ACER has been used.

The detection accuracy (Acc.) refers to the rate of correctly classified bona fide and PAs at $\delta = 0.5$:
\begin{equation}
\begin{split}
\mathrm{Acc}\left(\delta\right) &= \frac{1}{\text{\# samples}} \cdot  \Bigg\lbrace \left(1 - \mathrm{APCER}\left(\delta\right)\right) \cdot \left\lbrace \text{\# PA samples} \right\rbrace 
\\
&+  \left(1 - \mathrm{BPCER}\left(\delta\right)\right) \cdot \left\lbrace \text{\# BF samples} \right\rbrace \Bigg\rbrace 
\end{split}
\end{equation}

These will be used in Table~\ref{tab:sotaDL} where needed.

\begin{table*}[t]
\begin{small}
\begin{center}

\caption{Summary of the most relevant methodologies for fingerprint PAD based on non-conventional sensors.}\label{tab:sotaOther}
\centering
\begin{tabular}{YcRccc}
\toprule
\textbf{Year} & \textbf{Spectrum} & \textbf{Ref.} & \textbf{Description} & \textbf{Performance} & \textbf{Database (\# PAIs)}  \\
\midrule
\multirow{2}{*}{2008} &\multirow{2}{*}{430 -- 630 nm} & \multirow{2}{*}{\cite{Rowe-Lumidigm-WP-2008}}  & \multirow{2}{*}{Wavelet transform} & APCER = 0.9\% & Unavailable DB \\
 &  & & & BPCER = 0.5\% & (49)  \\ \cline{1-6}
\multirow{4}{*}{2011} & \multirow{2}{*}{400 -- 1650 nm} & \multirow{2}{*}{\cite{Hengfoss-spectroscopyPADfinger-FSI-2011}}  & \multirow{2}{*}{Spectroscopic properties} & \multirow{2}{*}{-} & Unavailable DB \\
 &  & & &  & (0)  \\ \cline{2-6}
 & OCT & \multirow{2}{*}{\cite{Chang-FpPANIR-InTech-2011}}  & \multirow{2}{*}{-} & \multirow{2}{*}{-} & Unavailable DB \\
 &400 -- 850 nm & & & & (-)  \\ \cline{1-6}
\multirow{6}{*}{2018} &  \multirow{4}{*}{1200 -- 1550 nm}  & \multirow{2}{*}{\cite{Gomez-Barrero-SWIR-SS-PAD-NISK-2018}}  & \multirow{2}{*}{Multi-spectral signatures} & APCER = 5.7\% & Unavailable DB \\
 &   & & & BPCER = 0.0\% & (12)  \\ \cline{3-6}
 & & \multirow{2}{*}{\cite{Tolosana-FingerPAD-CNN-SWIR-BIOSIG-2018}}  & \multirow{2}{*}{Pre-trained VGG19 model} & APCER = 0.0\% & Unavailable DB \\
 &  & & & BPCER = 0.0\% & (12)  \\ \cline{2-6}
  &  \multirow{2}{*}{1310 nm (LSCI)} & \multirow{2}{*}{\cite{Keilbach-PADlsciTexture-BIOSIG-2018}}  & \multirow{2}{*}{Texture descriptors} & APCER = 10.97\% & Unavailable DB \\
 &  & & & BPCER = 0.84\% & (32)  \\
\bottomrule
\end{tabular}\vspace{-0.5cm}
\end{center}
\end{small}
\end{table*}

\subsection{Non-Conventional Fingerprint Sensors}
\label{sec:related:nc}

To the best of our knowledge, the pioneering work on fingerprint multi-spectral PAD with non-conventional capacitive or optical sensors was carried out by Rowe \textit{et al.} in 
\cite{Rowe-Lumidigm-WP-2008}. The presented, and now widely used, Lumidigm sensor, captures multi-spectral images in four different wavelengths (i.e., 430, 530 and 630 nm, as well as white light). In their work, the authors studied the PAD capabilities of the combined images using absolute magnitudes of the responses of each image to dual-tree complex wavelets. In a self-acquired database including 49 PAI species, they obtained an APCER of 0.9\% for a BPCER of 0.5\%. Even if these results are remarkable, the PAD methods used are not described and not many details about the acquired database or the experimental protocol are presented. Therefore, it is difficult to establish a fair benchmark.

Three years later, Hengfoss \textit{et al.} analysed extensively the spectroscopic properties of living against the cadaver fingers using four wavelengths between 400 nm and 1630 nm \cite{Hengfoss-spectroscopyPADfinger-FSI-2011}. However, no PAIs were analysed in their work. Later that year, Chang \textit{et al.} studied in \cite{Chang-FpPANIR-InTech-2011} the complex properties of the skin, which differentiate it from PAIs, using optical coherence tomography (OCT) and nine different wavelengths between 400 nm and 850 nm. A single volunteer provided the bona fide and PA samples, and not many details about the algorithms used were reported.

More recently, in 2018, some preliminary PAD studies were carried out in \cite{Gomez-Barrero-SWIR-SS-PAD-NISK-2018,Tolosana-FingerPAD-CNN-SWIR-BIOSIG-2018} on a small database, comprising a total of 60 samples and 12 different PAI species, which was acquired at the University of South California within the BATL project \cite{BATL}. Gomez-Barrero \textit{et al.} extracted multi-spectral signatures from four different wavelengths in SWIR spectrum, achieving an APCER = 5.7\% and a BPCER = 0\%. In this case, all classification errors stem from a single PAI made with orange playdoh. In a subsequent work on the same database, Tolosana \textit{et al.} used a pre-trained VGG19 CNN model \cite{VGG19_2015} for PAD purposes. In this case, all 60 samples were correctly classified (i.e., APCER = BPCER = 0\%). 

Finally, Keilbach \textit{et al.} analysed in \cite{Keilbach-PADlsciTexture-BIOSIG-2018} the PAD capabilities of laser speckle contrast images (LSCI) over a larger database, also acquired within the BATL project and comprising 32 PAIs and more than 750 samples. In this case, several descriptors were extracted from the LSCI sequences, including the well-known local binary patterns (LBP) or the histogram of oriented gradients (HOG). The final cascaded score level fusion yielded an APCER = 10.97\% for a BPCER = 0.84\%. 

%
%
%

\begin{table*}[t]
\begin{small}
\begin{center}

\caption{Summary of the most relevant methodologies for fingerprint PAD based on DL approaches.}\label{tab:sotaDL}
\centering
\begin{tabular}{lYRlcc}
\toprule
\textbf{Category} & \textbf{Year} & \textbf{Ref.} & \textbf{Description} & \textbf{Performance} & \textbf{Database (\# PAIs)} \\ \midrule
\multirow{8}{*}{Full Sample} & \multirow{2}{*}{2015} & \multirow{2}{*}{\cite{Menotti-PADdeepRep-TIFS-2015}} & \multirow{2}{*}{CNN optimization} &\multirow{2}{*}{Acc. = 98.97\%} & LivDet 2013 \\ 
 & & & & & (7) \\ \cline{2-6}
 & \multirow{6}{*}{2016} & \multirow{2}{*}{\cite{Nogueira-PADfingerprintCNN-TIFS-2016}} & Pre-trained CNNs &\multirow{2}{*}{ACER = 2.90\%} & LivDet 2009-13  \\
 & & & (Best: VGG) & & (8) \\ \cline{3-6}
  & & \multirow{2}{*}{\cite{Kim-fingerprintPAD-DNN-PRL-2016}} & \multirow{2}{*}{DBN with RBMs} &\multirow{2}{*}{Acc. = 97.10\%} & LivDet 2013  \\
 & & & & & (7) \\ \cline{3-6}
   & & \multirow{2}{*}{\cite{Marasco-PADfingerprintCNN-2016-HST}} & Pre-trained CNNs and Siamese networks &\multirow{2}{*}{Acc. = 96.60\%} & LivDet 2011-13  \\
 & & & (Best: GoogLeNet) & & (8) \\ \cline{1-6}
  \multirow{2}{*}{ROI} & \multirow{2}{*}{2017} & \multirow{2}{*}{\cite{Yuan-FingerprintPAD-CNN-PCA-CMC-2017}} & CNNs + ROI and PCA optimization &ACER = 4.57\% (2011) & LivDet 2011-13  \\
 & & & and SVM classification & ACER = 7.25\% (2013) & (8) \\ \cline{1-6}
\multirow{16}{*}{Patch-wise}   & \multirow{2}{*}{2015} & \multirow{2}{*}{\cite{Wang-FpPAD-DCNN-CCBR-2015}} & \multirow{2}{*}{DCNN (CiFar10-Net + FingerNet)} & ACER = 0.88\% (2011) & LivDet 2011-13 \\ 
 & & & & ACER = 0.90\% (2013) & (8) \\ \cline{2-6}
 & \multirow{2}{*}{2016} & \multirow{2}{*}{\cite{Park-PADfpCNN-BIOSIG-2016}} & \multirow{2}{*}{CNN trained from scratch} &  \multirow{2}{*}{ACER = 3.42\%} & LivDet 2009 \\ 
 & & & & & (Identix, 3) \\ \cline{2-6}
 & \multirow{8}{*}{2017} & \multirow{2}{*}{\cite{Jang-fingerprintPADcontrastCNN-ICISA-2017}} & Contrast enhancement &\multirow{2}{*}{ACER = 0.20\%} & ATVS FP  \\
 &  & & + Ad hoc CNN & & (2) \\ \cline{3-6}
   & & \multirow{2}{*}{\cite{Souza-FingerprintPAD-DBM-IJCNN-2017}} & \multirow{2}{*}{Deep Boltzmann Machine} &\multirow{2}{*}{Acc. = 85.96\%} & LivDet 2013  \\
 &  & &  & & (7) \\ \cline{3-6}
  & & \multirow{2}{*}{\cite{Toosi-fingerprintPADpatchCNN-ICCI-2017}} & Pre-trained AlexNet & ACER = 4.63\% (2011) & LivDet 2011-13   \\ 
 &  &  &+ Data augmentation and log-likelihood& ACER = 1.90\% (2013) & (8) \\  \cline{3-6}
 & & \multirow{2}{*}{\cite{Pala-FingerprintPAD-DeepTriplet-ICIP-2017}} & \multirow{2}{*}{Deep triplet embedding} &\multirow{2}{*}{ACER = 1.74\%} & LivDet 2009-13  \\
 &  & &  & & (8) \\ \cline{2-6}
 & \multirow{4}{*}{2018} & \multirow{2}{*}{\cite{chugh-fingerprintPADcnn-TIFS-2018}} & Pre-trained MobileNet & ACER = 0.96\% & LivDet 2011-15 (11)   \\
 &  &  & + Minutiae patches & ACER = 2.00\% & Own DB (12) \\ \cline{3-6}
  & & \multirow{2}{*}{\cite{Park-FingerprintPADCNN-arxiv-2018}} & Fully CNN (SqueezeNet)& \multirow{2}{*}{ACER = 1.43\%} & LivDet 2011-15   \\
 &  &  & + Data augmentation &  & (11) \\ \cline{1-6}
   \multirow{2}{*}{Deep Fusion} & \multirow{2}{*}{2017} & \multirow{2}{*}{\cite{Toosi-2017-FingerprintPAD-FeatureFusion-Access-2017}} & Texture based features & \multirow{2}{*}{ACER $\approx$1.70\%}  & LivDet 2009-13  \\
 & & & and DNN fusion & & (8) \\ 
\bottomrule
\end{tabular}\vspace{-0.5cm}
\end{center}
\end{small}
\end{table*}

\subsection{Deep Learning for Conventional Sensors}
\label{sec:related:dl}

The DL based fingerprint PAD proposed in the literature can be widely classified depending on the input of the networks into: $i)$ using the full samples as input to the network, $ii)$ cropping the region of interest (ROI) and feeding it to the network, and $iii)$ extracting patches from the ROI as input. Moreover, some articles $iv)$ use the network for feature level fusion of handcrafted descriptors. In the following, the main studies in all categories are summarised. 

\textbf{Full samples}. To the best of our knowledge, the first work on fingerprint PAD based on deep learning algorithms was presented in 2015 by Menotti \textit{et al.} \cite{Menotti-PADdeepRep-TIFS-2015}. The authors proposed two different CNN optimization approaches for the particular purpose of PAD. On the one hand, the architecture was optimized with feedforward convolutional operations and hyperparameter optimization. On the other hand, the inner weights of the network were optimized via back-propagation. Both techniques were tested on iris, face and fingerprint benchmarks, thus proving the generalisation capabilities of the proposal. Their best fingerprint related results achieved an average detection accuracy, Acc., across the four fingerprint sensors of LivDet 2013 of 98.97\%. 

A year later, three different approaches were proposed. Nogueira \textit{et al.} \cite{Nogueira-PADfingerprintCNN-TIFS-2016} tested three different CNNs, namely: $i)$ the pre-trained VGG \cite{VGG19_2015}, $ii)$ the pre-trained Alexnet \cite{Alexnet-ImageClassificationWithDeepCNN-ANIPS-2012}, and $iii)$ a CNN with randomly initialised weights and trained from scratch. They compared the ACER obtained with the networks over the LivDet 2009, 2011 and 2013 databases to a classical state-of-the-art algorithm based on LBP. In the evaluation, the best detection performance is achieved using a VGG pre-trained model and data augmentation (average ACER = 2.9\%), with a clear improvement with respect to LBP (average ACER = 9.6\%). It should be also noted, that the ACER decreased between 25\% and 50\% (relative decrease) for all three networks tested when data augmentation was used.

Then, Kim \textit{et al.} analysed the use of deep belief networks based on superimposed restricted Boltzmann machines (RBMs) \cite{Kim-fingerprintPAD-DNN-PRL-2016}. The global network is trained in a two-stage manner with layer-wise greedy training and fine-tuning with labelled inputs. On LivDet 2013, they achieved a detection accuracy Acc. of 97.10\%, noting again the considerable enhancement achieved with data augmentation.

Also Marasco \textit{et al.} explored in \cite{Marasco-PADfingerprintCNN-2016-HST} two different pre-trained CNNs: $i)$ CaffeNet \cite{Alexnet-ImageClassificationWithDeepCNN-ANIPS-2012}, and $ii)$ GoogLeNet \cite{Szegedy-Googlenet-CNNs-CVPR-2015}. Furthermore, the performance of such networks was compared to a Siamese network, which optimised a metric distance to yield high bona fide - PA distances and low bona fide - bona fide distances. In a thorough evaluation on LivDet 2011 and 2013, a detection accuracy over 96\% was achieved for GoogLeNet, closely followed by the other networks. The authors showed an accuracy decrease when dealing with either unknown attack or a cross-sensor scenario.

\textbf{ROI}. In 2017, Yuan \textit{et al.} followed a different approach to optimise the performance of CNN models \cite{Yuan-FingerprintPAD-CNN-PCA-CMC-2017}. First, only the ROI was fed to the network. Then, principal component analysis (PCA) was introduced for each convolutional and pooling operation in order to discard non-relevant information. Finally, the output was classified with SVMs. This way, no data augmentation was required to achieve a 4.57\% ACER over LivDet 2013, thereby outperforming other existing approaches.

\textbf{Patch-wise}. In 2015, a different two-step approach was proposed by Wang \textit{et al.} \cite{Wang-FpPAD-DCNN-CCBR-2015}. First, the ROI of the fingerprint was segmented. Then, two deep CNNs (DCCNs) were used in a patch-wise manner: $i)$ the CiFar10-Net \cite{Wan-ciFar10Net-ICML-2013}, and $ii)$ the self-developed Finger-Net, yielding an ACER under 1\% over LivDet 2011 and 2013.

In 2016, Park \textit{et al.} extracted random patches from the fingerprint samples and trained a CNN from scratch in \cite{Park-PADfpCNN-BIOSIG-2016}, achieving an ACER = 3.4\% over the Identix subset of LiveDet 2009. 

In 2017, Jang \textit{et al.} proposed contrast enhancement and block-wise processing of the fingerprint to improve the state of the art results achieved with DL \cite{Jang-fingerprintPADcontrastCNN-ICISA-2017}. The blocks were then combined with a majority voting rule. They also designed a CNN from scratch inspired in the VGG19 model, and evaluated the proposed approach over the ATVS fake fingerprint DB \cite{Galbally-FingerprintPAEval-TS-2011}. An ACER of 0.2\% was reported.

Souza \textit{et al.} analysed again in \cite{Souza-FingerprintPAD-DBM-IJCNN-2017} the use of Boltzmann machines, this time in a patch-wise manner and using a majority vote rule. In particular, they used deep Boltzmann machines (DBMs), which can learn more complex and internal representations from a low number of labelled samples. The accuracy obtained over LivDet 2013 was 85.96\%.

Following this patch-wise trend, Toosi \textit{et al.} tested in \cite{Toosi-fingerprintPADpatchCNN-ICCI-2017} the accuracy of AlexNet with data augmentation. For classification, the scores are calibrated using log-likelihood ratios. The average ACER on LivDet 2011 and 2013 is 3.26\%.

Similarly, Pala \textit{et al.} tested the feasibility of using deep triplet embedding for PAD purposes \cite{Pala-FingerprintPAD-DeepTriplet-ICIP-2017}. In contrast to Siamese networks, this method requires no enrolment database, since the triplets are selected from patches within the input sample. Over LivDet 2009 to 2013, an ACER of 1.74\% was reported. The robustness to unknown attacks was also evaluated on LivDet 2013, achieving ACERs much lower than other approaches (e.g., 0.7\% vs 1.4\% for Siamese networks for the modasil PAIs).

\begin{figure*}[t]
\centering
 \centerline{\includegraphics[width=.65\linewidth]{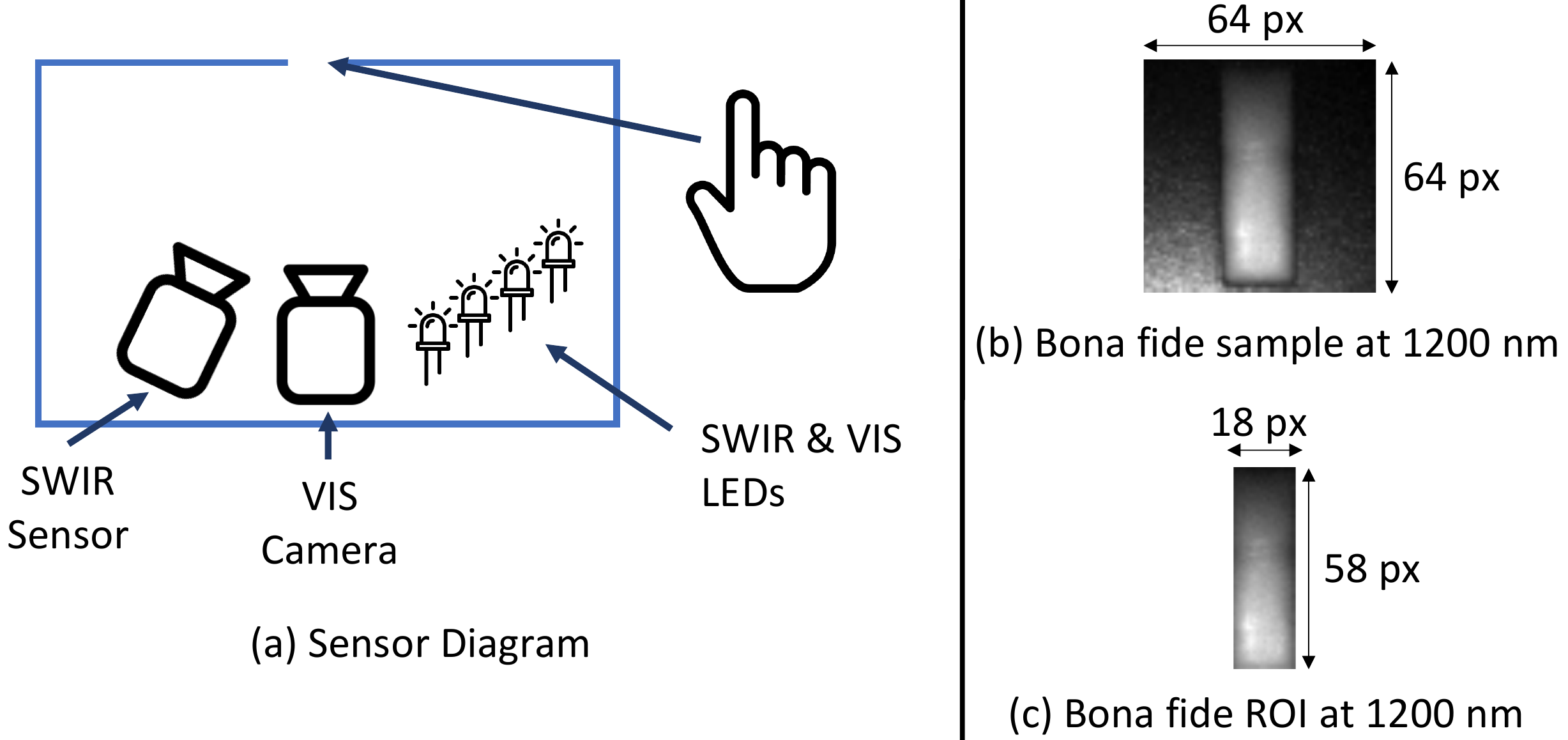}}
 \caption{Finger sensor diagram. \textbf{Left}: a diagram of the inner components: two different sensors for the SWIR images and the visible (VIS) light images, together with the corresponding LEDs. \textbf{Right}: a sample and the corresponding ROI for a bona fide at 1200 nm.} \label{fig:sensor}\vspace{-0.4cm}
\end{figure*}

In 2018, Chugh \textit{et al.} presented in \cite{chugh-fingerprintPADcnn-TIFS-2018} a different way to extract fingerprint patches: around the minutiae. The idea behind this patch computation is the fact that PAIs can present spurious minutiae, which can be surrounded by a distinct texture. Therefore, these patches were fed to the MobileNet pre-trained network \cite{DBLP:journals/corr/HowardZCKWWAA17}. The detection performance was evaluated on LivDet 2011 to 2015, achieving a remarkable ACER of 0.96\% on average. However, the ACER increased to 2.0\% for a self-acquired database, comprising a larger number of PAIs (12).

In the same year, Park \textit{et al.} developed in \cite{Park-FingerprintPADCNN-arxiv-2018} a fully CNN based on the Fire module of SqueezeNet \cite{Iandola-SqueezeNet-arxiv-2016}. They analysed different patch sizes and compared the common voting method to an optimal thresholding approach, which yielded a better performance: an ACER of 1.43\% over LivDet 2011 to 2015.

\textbf{Deep fusion}. Toosi \textit{et al.} proposed in \cite{Toosi-2017-FingerprintPAD-FeatureFusion-Access-2017} a completely different approach to use DL for fingerprint PAD. Instead of using deep networks for feature extraction, ten different hand-crafted descriptors, including the well-known local phase quantization (LPQ), binary statistical features (BSIF) or scale invariant feature transform (SIFT) were fed to a self-developed deep network (Spidernet) for final fusion and classification. The performance was compared to classical fusion approaches, such as SVMs and AdaBoost, and ACER around 1.6-1.8\% were reported for LivDet 2009 to 2013.





\section{Presentation Attack Detection Methodology: Hardware}
\label{sec:swir}

The finger SWIR capture device used for the present work was developed within the BATL project \cite{BATL} in cooperation with our project partners. A general diagram of its inner components is included in Fig.~\ref{fig:sensor} (a). As it may be observed, the camera and lens are placed inside a closed box, which includes an open slot on the top. When the finger is placed there, all ambient light is blocked and therefore only the desired wavelengths are used for the acquisition. In particular, we have used a Hamamatsu InGaAs SWIR sensor array, which captures $64 \times 64$ px images, with a 25 mm fixed focal length lens optimised for wavelengths within 900 -- 1700 nm. More specifically, the following SWIR wavelengths were selected for PAD purposes: $\lambda_1 = 1200$ nm, $\lambda_2 = 1300$ nm, $\lambda_3 = 1450$ nm, and $\lambda_4 = 1550$ nm. These are similar to the wavelengths considered in \cite{Steiner-facePADswir-Sensors-2016} for the skin vs.\ non-skin facial classification. 

An example of the acquired images for a bona fide sample is shown in Fig.~\ref{fig:sensor} (b) for the 1200 nm wavelength. As it may be observed, before applying any PAD algorithm, the region of interest (ROI) (i.e., the central finger-slot region corresponding to the open slot where the finger is placed) needs to be extracted from the background. Given that the finger is always placed over the fixed open slot, and the camera does not move, a simple fixed size cropping can be applied. The ROI corresponding to Fig.~\ref{fig:sensor} (b) with a size of $18 \times 58$ px is depicted in Fig.~\ref{fig:sensor} (c).

Finally, the four samples acquired from two bona fides (a, b) and three PAIs (c to e) fabricated with different materials are included in Fig.~\ref{fig:input_DNN}: (c) a full yellow playdoh finger, (d) a monster latex overlay, and (e) a glue overlay. As it may be observed, the playdoh finger shows some similarities with respect to the bona fide presentations (i.e., a similar change of intensity across wavelengths), which will make the PAD task harder. However, the change trend is completely different for the other two PAIs, thereby making it easier to discriminate them from bona fide presentations.


In addition to the SWIR images captured by the device, fingerprint verification can be carried out with contactless finger photos acquired in the visible spectrum with a 1.3 MP camera and a 35 mm VIS-NIR lens, which are placed next to the SWIR sensor within the closed box (see Fig.~\ref{fig:sensor} (a)). Note that the project sponsor IARPA has indicate that they make the SWIR finger database available in the near future such that research results presented in this paper can be reproduced. 


\section{Presentation Attack Detection Methodology: Software}
\label{sec:pad}

This section describes the state-of-the-art software methods proposed in order to detect fingerprint PAs, as summarised in Fig.~\ref{fig:diagram}. Two different approaches are studied: $i)$ handcrafted features, and $ii)$ deep learning features. For both approaches, the information provided by the sensor described in Sect.~\ref{sec:swir} is used as input.

In general, it should be noted that each individual score $s_i$ generated by the individual PAD algorithms needs to be transformed into a single range to allow the final fusion and a fair benchmark. In compliance with the ISO/IEC 30107-2 standard on biometric presentation aattack detection -- Part 2: data formats \cite{ISO-IEC-30107-2-PAD-data-format-170328}, we define $s_i \in [0, 100]$, where low values close to 0 will represent bona fide presentations and high values close to 100 will denote presentation attacks.

\subsection{Handcrafted Features}
\label{sec:pad:svm}

As it was firstly proposed in \cite{Gomez-Barrero-SWIR-SS-PAD-NISK-2018}, this method builds upon the spectral signatures of the pixels across all four acquired wavelengths, in order to capture the different properties attributed to skin (i.e., bona fide presentation) and non-skin (i.e., PAI) materials. In particular, let us define the spectral signature $\mathbf{ss}$ of a pixel with coordinates $\left( x, y\right)$ as follows:
\begin{equation}
\mathbf{ss}\left(x, y\right) = \left(i_1, \dots, i_N\right)
\end{equation}
where $i_n$ represents the intensity value of the pixel for the $n$th wavelength $\lambda_n$. In our particular case study, $N = 4$. 

However, such a representation is vulnerable to illumination changes. Even if they have been minimised in the sensor due to only having the finger slot open to the outer world, thinner fingers for instance can let some tiny amounts of light through. As a consequence, in order to achieve a signature independent of the absolute brightness of the image at hand, a normalised signature is computed. In addition, since our final goal is to capture the distinct trends across different wavelengths shown in Fig.~\ref{fig:input_DNN} for the bona fides and the PAIs, only differences between wavelengths will be used as final handcrafted features. Therefore, the final normalised difference vector $\mathbf{d}\left( x, y\right) $ of one pixel is computed as follows:
\begin{eqnarray}
d\left[i_a, i_b\right] &=& \frac{i_a - i_b}{i_a + i_b}
\\
\mathbf{d}\left( x, y\right) &=& \left\lbrace d\left[ i_a, i_b\right] \right\rbrace_{a, b \le N, a  \ne b}
\end{eqnarray}
with $ -1 \le d\left[i_a, i_b\right]  \le 1$. In other words, the normalised differences between all possible wavelength combinations are computed. For our case study with $N = 4$, a total number of six differences are calculated. These normalised difference vectors $\mathbf{d}\left(x, y\right)$ will be used to classify skin vs.\ non-skin pixels using an SVM. 

The procedure so far performs a pixel wise classification. Hence, the final score $s_\text{ss}$ returned by the PAD method will be the proportion of non-skin pixels of the sample ROI in a range of 0 to 100.


\subsection{Deep Learning Features}
\label{sec:pad:DL_features}

\begin{figure}[t]
\centering
 \centerline{\includegraphics[width=0.80\linewidth]{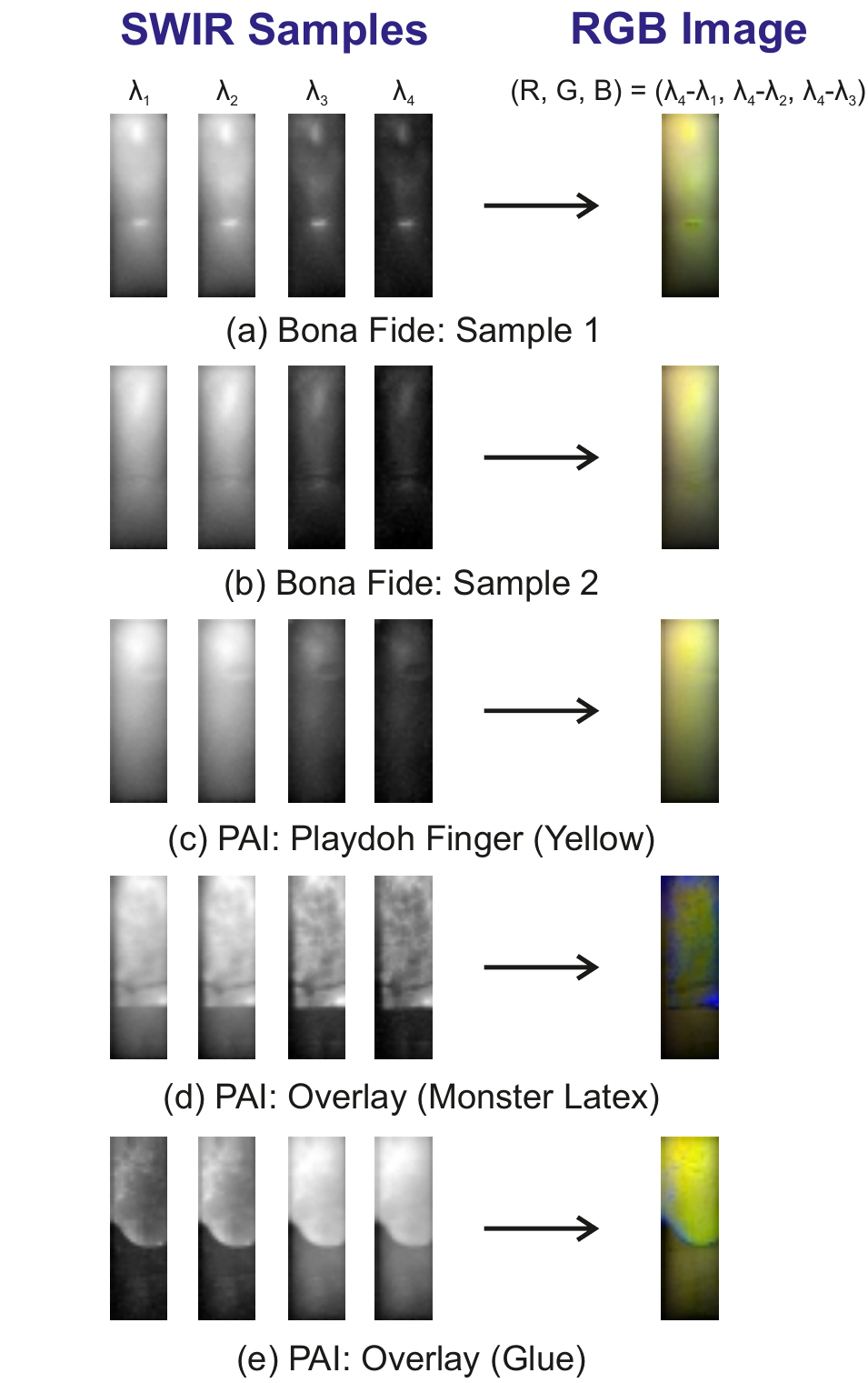}}
 \caption{Examples of bona fides and PAs acquired by the SWIR sensor and the final RGB image created for the input of the deep neural network systems (see Eq.~\ref{eq:RGB}).} \label{fig:input_DNN}\vspace{-0.4cm}
\end{figure}

CNNs have been one of the most successful deep neural network architectures in the last years. Some of their key design principles were drawn from the findings of the Neurophysiologists Nobel Prizes David Hubel and Torsten Wiesel in the field of human vision \cite{Goodfellow-et-al-2016}. Traditional (a.k.a. plain) CNN based systems are mainly composed of convolutional and pooling layers. The former extracts patterns from the images through the application of several convolutions in parallel to local regions of the images. These convolutional operations are carried out by means of different kernels, adapted by the learning algorithm, and which assign a weight to each pixel of the local region of the image depending on the type of patterns to be extracted. Therefore, each kernel of one convolutional layer is focused on extracting different patterns, such as horizontal or vertical edges, over image patches whose size is determined by the dimension of the layer. The output of these operations produces a set of linear activations (a.k.a. feature map), which serve as input to nonlinear activations, such as the rectified linear activation function (ReLU). Finally, it is common to use pooling layers to make the representation invariant to small translations of the input. The pooling function replaces the output of the network at a certain region with a statistical summary of the nearby outputs, and facilitates the learning convergence. For instance, the max-pooling function selects the maximum value of the region.

As it was summarised in Fig.~\ref{fig:diagram}, in this study we explore the potential of deep learning features in comparison to handcrafted features by means of two different strategies: $i)$ using CNNs as an end-to-end approach (i.e., for both feature extraction and classification), and $ii)$ using CNNs as feature extractors in combination with SVMs for classification. In addition, two different training scenarios have been analysed, namely: $i)$ training CNN models from scratch, and $ii)$ adapting CNN pre-trained models.

For the input of the networks, and in order to consider the information provided by the four wavelengths captured by the sensor, we need to build a single RGB image. To that end, each dimension or channel of the RGB space will comprise information stemming from different SWIR wavelengths or combinations thereof. To maximise the discriminative power of the input images, we analysed which wavelengths provided a higher inter-class (i.e., between bona fide and PA presentations) and a lower intra-class (i.e., within the bona fide presentation samples) variation in terms of the heatmaps of the differences between samples. That is, to estimate the inter-class variability we computed the pixel wise difference of bona fide and PA samples, and for the intra-class variability, the differences between bona fide samples. The former should have high intensity values and the latter low values. After an exhaustive analysis of the different possible combinations, we defined the three dimensions as follows:
\begin{equation}
\mathbf{image}\left(R, G, B\right) = (\lambda_4 - \lambda_1, \lambda_4 - \lambda_2, \lambda_4 - \lambda_3)\label{eq:RGB}
\end{equation}

Fig.~\ref{fig:input_DNN} shows examples of bona fides and PAIs acquired by the SWIR sensor and the final RGB image created following Eq.~\ref{eq:RGB}. This RGB image will serve as input for the deep neural network systems. All strategies have been implemented under the Keras framework using Tensorflow as back-end, with a NVIDIA GeForce GTX 1080 GPU. Adam optimizer is considered with a learning rate value of 0.0001 and a loss function based on binary cross-entropy. We now describe the details of each of the deep learning strategies analysed in this work.

\vspace*{0.2cm}

\subsubsection{\textbf{Training CNN Models from Scratch}}
\label{sec:pad:dl}

The first approach is focused on training \textbf{residual CNNs} \cite{DBLP:journals/corr/HeZRS15} from scratch. These networks have outperformed traditional (a.k.a. plain) networks in many different datasets such as ImageNet 2012 \cite{DBLP:journals/corr/RussakovskyDSKSMHKKBBF14}, CIFAR-10 \cite{CIFAR10}, PASCAL VOC 2007/2012 \cite{everingham2010pascal} and COCO \cite{DBLP:journals/corr/LinMBHPRDZ14} for both image classification and object detection tasks. The peculiarity of this network is the insertion of shortcut connections every few stacked layers, converting the plain network into its residual version. This allows to use deeper neural network architectures as well as accelerating the training of the networks significantly \cite{DBLP:journals/corr/HeZRS15, DBLP:journals/corr/SzegedyIV16}.

\begin{figure}[t]
\centering
\centerline{\includegraphics[width=\linewidth]{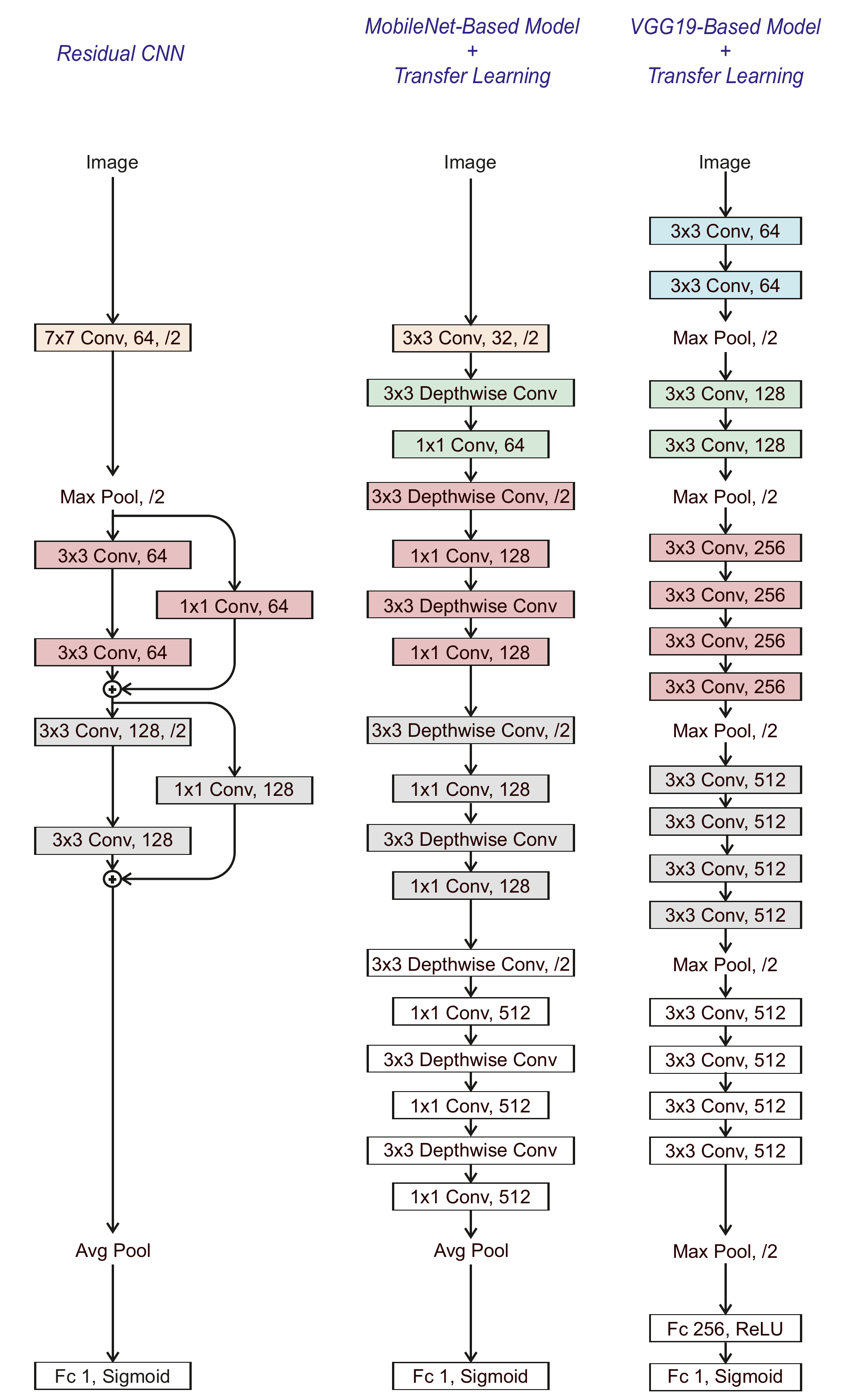}}
\caption{Proposed network architectures. \textbf{Left:} the residual CNN trained from scratch using only the SWIR fingerprint database (319,937 parameters). \textbf{Middle:} the pre-trained MobileNet-based model (815,809 parameters). \textbf{Right:} the pre-trained VGG-19-based model (20,155,969 parameters). Both middle and right networks are adapted using transfer learning techniques over the last white-background layers} \label{fig:network_architectures}\vspace{-0.4cm}
\end{figure}

Our proposed residual CNN is depicted in Fig.~\ref{fig:network_architectures} (left). Batch normalization (BN) is applied right after each convolution and before the activation following \cite{DBLP:journals/corr/IoffeS15}. All activation functions are based on ReLU apart from the Sigmoid activation used in the last fully-connected layer, which provides output scores between 0 and 100.

\vspace*{0.2cm}

\subsubsection{\textbf{Adapting Pre-Trained CNN Models}}
\label{sec:pad:tl}
The second approach evaluates the potential of state-of-the-art pre-trained models for fingerprint PAD. In order to adapt the pre-trained models to our task, we replace and retrain the classifier (i.e., the fully-connected layers), and adapt the weights of the last convolutional layers to the fingerprint PAD task. The reason for adapting only the last convolutional layers lies on the fact that the first layers of the CNN extract more general features related to directional edges and colours, whereas the last layers of the network are in charge of extracting more abstract features related to the specific task. We propose to use both MobileNet and VGG-19 network architectures pre-trained using the ImageNet database \cite{DBLP:journals/corr/HowardZCKWWAA17, VGG19_2015}. This database contains more than one million images from 1000 different classes, thereby allowing the extraction of very robust features in the first layers \cite{DBLP:journals/corr/RussakovskyDSKSMHKKBBF14}.



Fig.~\ref{fig:network_architectures} (middle) shows the architecture of our adapted \textbf{MobileNet} network. This architecture has been modified compared to the original version by removing some of the last convolutional layers in order to reduce the complexity of the features extracted. Furthermore, the fully-connected layers designed for the ImageNet classification task have been also removed. This network is based on depthwise separable convolutions, which factorize a standard convolution into: $i)$ a depthwise convolution, and $ii)$ a $1x1$ convolution called pointwise convolution. Therefore, the depthwise convolution applies a single filter to each input channel, and the pointwise convolution subsequently applies a $1x1$ convolution to combine the outputs of the depthwise convolution \cite{DBLP:journals/corr/HowardZCKWWAA17}. Downsampling is directly applied by the convolutional layers that have a stride of 2 (represented by /2 in the convolutional layers of Fig.~\ref{fig:network_architectures}). This network architecture allows to reduce both model size and training/testing times, thus being a good solution for mobile and embedded vision applications. It has been tested in different datasets such as ImageNet \cite{DBLP:journals/corr/RussakovskyDSKSMHKKBBF14}, PlaNet \cite{DBLP:journals/corr/WeyandKP16} and COCO \cite{DBLP:journals/corr/LinMBHPRDZ14} with state-of-the-art results.

Finally, Fig.~\ref{fig:network_architectures} (right) shows the architecture of the adapted \textbf{VGG-19} network \cite{VGG19_2015}. This architecture has also been modified replacing the last 3 fully-connected layers with 2 fully-connected layers (with a final sigmoid activation). This network architecture belongs to the family of traditional or plain networks and appeared before the residual and MobileNet configurations. Despite of that, and due to its simplicity, it is one of the most used network architectures nowadays providing very good results in many different competitions. 


\vspace*{0.2cm}

\subsubsection{\textbf{Using CNNs as Feature Extractors}}
\label{sec:pad:dlsvm}

In addition to the end-to-end approaches described in Sect.~\ref{sec:pad:dl} and~\ref{sec:pad:tl}, we also analyse the potential of adapting and using all the aforementioned CNNs (i.e., the residual CNN trained from scratch, the adapted MobileNet CNN and the adapted VGG-19 CNN) as feature extractors. For this strategy, we consider the same architecture networks described in Fig.~\ref{fig:network_architectures}, but removing the last fully-connected layers in order to use only the features provided by the last convolutional layer (after the Average or Max pool layers, respectively). Then, these features are transformed to the range $[0, 1]$ and subsequently used to train an SVM for final classification purposes.

\subsection{Fused Approach}
\label{sec:pad:fusion}

Finally, we analyse to which extent the proposed algorithms complement each other to enhance the final fingerprint PAD decisions. To that end, the algorithms are fused with a weighted sum of the individual PAD scores as follows:
\begin{equation}
s = \left(1 - \alpha\right) \cdot s_1 + \alpha \cdot s_2\label{eq:fusion}
\end{equation}
where $s_i$ with $i \in \left\lbrace \text{ss}, \text{res}, \text{mob}, \text{VGG} \right\rbrace$ represent the individual scores output by the approaches described above, $\alpha$ the fusion weight, and $s$ the final fusion score.

\section{Experimental Framework}
\label{sec:setup}

\begin{table}[t]
\caption{PAI species included in the experimental work of this study. PAI species used only for testing and not for training (i.e., unknown attacks) have been underlined.}\label{table:types_PAIs}
\begin{adjustbox}{width=0.49\textwidth}
\begin{tabular}{ll}
\hline
\multicolumn{1}{l}{\textbf{Type}} & \textbf{Description}                                                              \\ \hline
\multicolumn{1}{l}{Dragon Skin}   & Finger, conductive, conductive nanotips white, \underline{graphite}                                    \\ 
\multicolumn{1}{l}{Latex}         & Finger                                                                             \\ 
\multicolumn{1}{l}{Overlay}       & Conductive silicone, monster latex, glue, silicone, \underline{urethane}, wax, dragon skin     \\ 
\multicolumn{1}{l}{Playdoh}       & Black, blue, green, orange, pink, purple, red, teal, \underline{yellow}                \\ 
\multicolumn{1}{l}{Printed}       & 2D photograph/matte paper,  3D normal/Ag paint,                              \\ 
\multicolumn{1}{l}{Silicone}      & Barepaint coating, finger flesh/yellow, graphite, normal, \underline{coating} \\ 
\multicolumn{1}{l}{Silly Putty}   & Glow in the dark, normal, \underline{metallic}                                                 \\ 
\multicolumn{1}{l}{Wax}           & Finger \\ \hline                                                                            
\end{tabular}
\end{adjustbox}\vspace*{-0.3cm}
\end{table}

\subsection{Database}
\label{sec:setup:db}


The database considered in the experimental evaluation was acquired within the BATL research project \cite{BATL} in collaboration with our partners at the Univiersity of South California (USC). The project is financed by IARPA ODIN program \cite{odinThorProgram}. Data were collected in two different stages and comprise both bona fide and PA samples.

For the bona fide samples, a total of 163 subjects participated during the first stage. For each of them, all 5 fingers of the right hand were captured. For the second stage, there were a total of 399 subjects. Index, middle and ring fingers of both hands were captured from each subject. It is important to highlight that people from different gender, ethnicity and age were considered during the acquisition, in order to evaluate the systems and algorithms in realistic conditions.

For the PA samples, the selection of the PAI fabrication materials was based on the requirements of IARPA ODIN program evaluation, covering the most challenging PAIs \cite{Sousedik-PAD-Survey-IET-BMT-2014, Marasco-PAD-SurveyFingerprint-CSUR-2015}. There are a total of 35 different PAI species, which can be further categorized into eight main groups, namely: dragon skin, latex, overlay, playdoh, printed fingers, silicone, silly putty and wax. All details are included in Table~\ref{table:types_PAIs}. 

Finally, all captured samples were manually reviewed in order to remove all samples with operational errors (e.g., finger movement) or hardware failures, ending up with a total of 4,290 and 443 bona fide and PA samples, respectively.

%
%
%
%
%
%
%

\subsection{Experimental Protocol}
\label{sec:setup:prot}

The main goal behind the experimental protocol design was to analyse and prove the soundness of our proposed fingerprint PAD approach in a realistic scenario. Therefore, the database described in the previos section is split into non-overlapping training, validation and test datasets, following the same procedure considered in previous works \cite{VGG19_2015, DBLP:journals/corr/HeZRS15}. All details are shown in Table~\ref{tab:partition_datasets}. In order to allow a fair benchmark among the approaches described in Sect.~\ref{sec:pad}, the same partitions will be used for all the experiments.

\begin{table}[t]
\small
	\centering
    	\caption{\label{tab:partition_datasets}Partition of training, validation and test datasets.}\vspace*{-0.2cm}
	\begin{tabular}{lccc}
		\toprule
		 & \# Samples & \# PA Samples & \# BF Samples\\
		\midrule
		Training set & 260 & 130 & 130 \\
		Validation set & 180 & 90 & 90 \\
		Test set & 4293 & 222 & 4071 \\
		\bottomrule
	\end{tabular}\vspace*{-0.3cm}
\end{table}

\begin{figure*}[tb]
\centering
\begin{subfigure}[tb]{0.32\textwidth}
\centering
\centerline{\includegraphics[width=\linewidth]{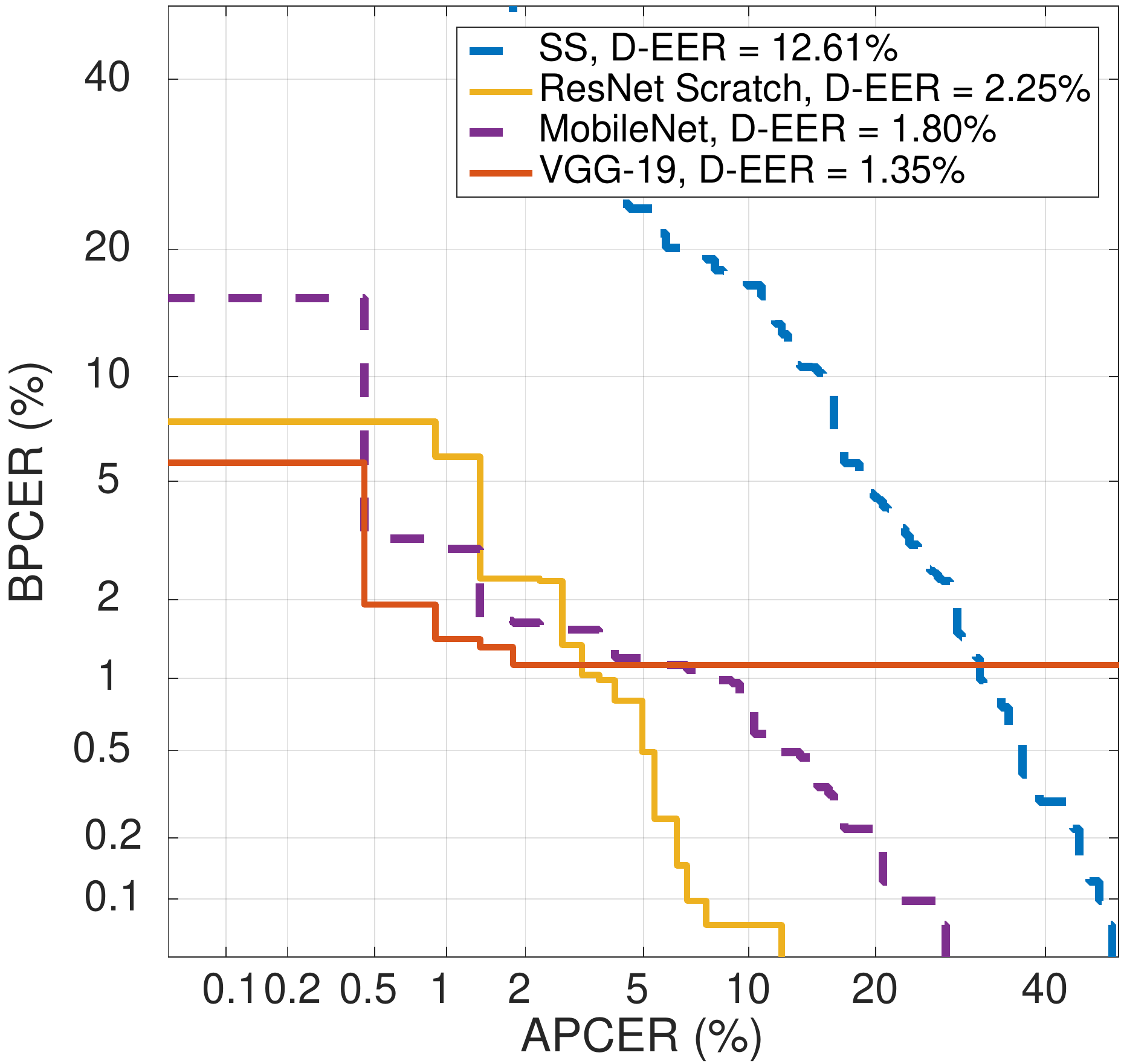}}
\caption{} \label{fig:DET_results}
\end{subfigure}
\begin{subfigure}[tb]{0.32\textwidth}
\centerline{\includegraphics[width=\linewidth]{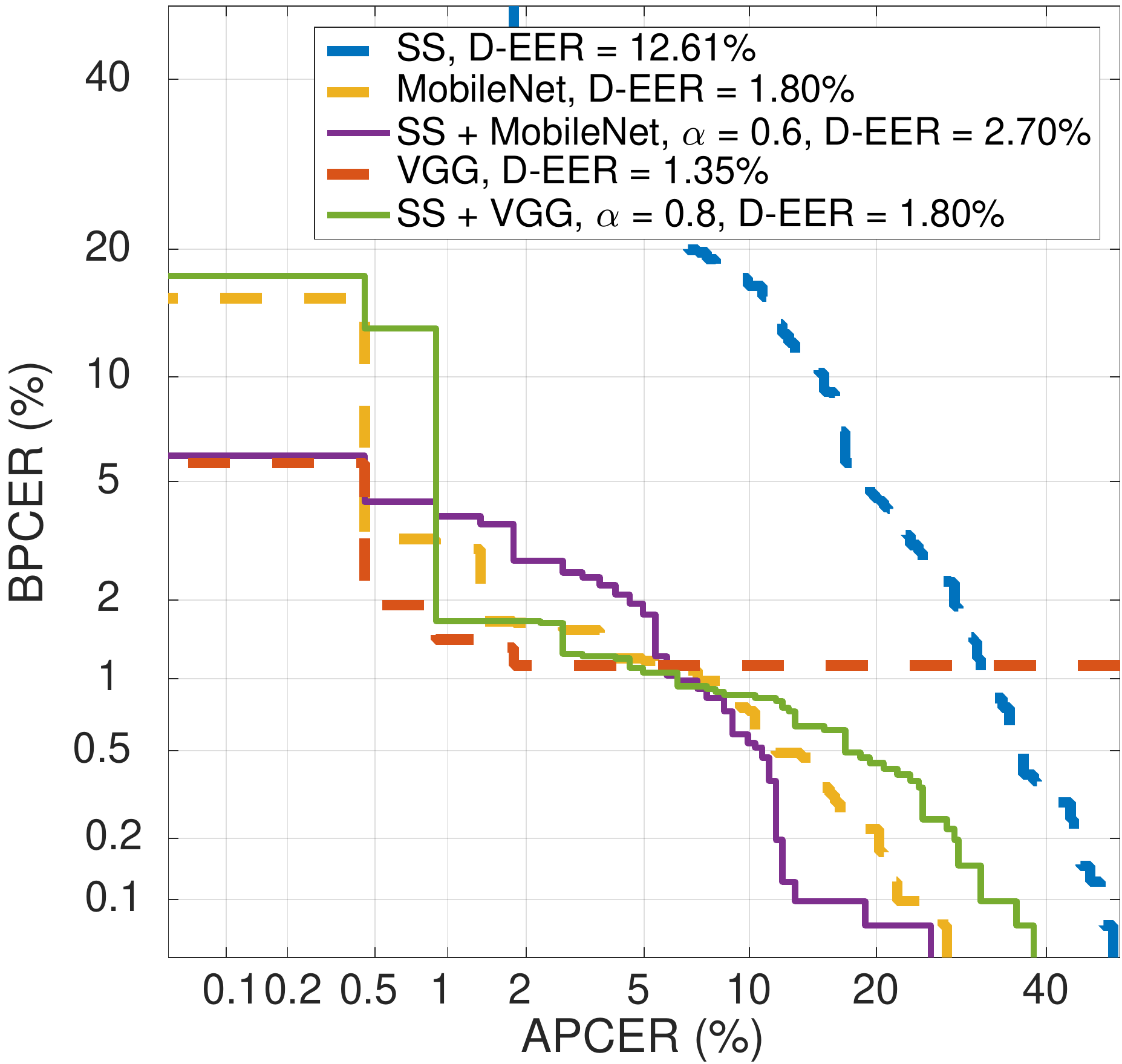}}
\caption{} \label{fig:DET_SSfusion}
\end{subfigure}
\begin{subfigure}[tb]{0.32\textwidth}
\centerline{\includegraphics[width=\linewidth]{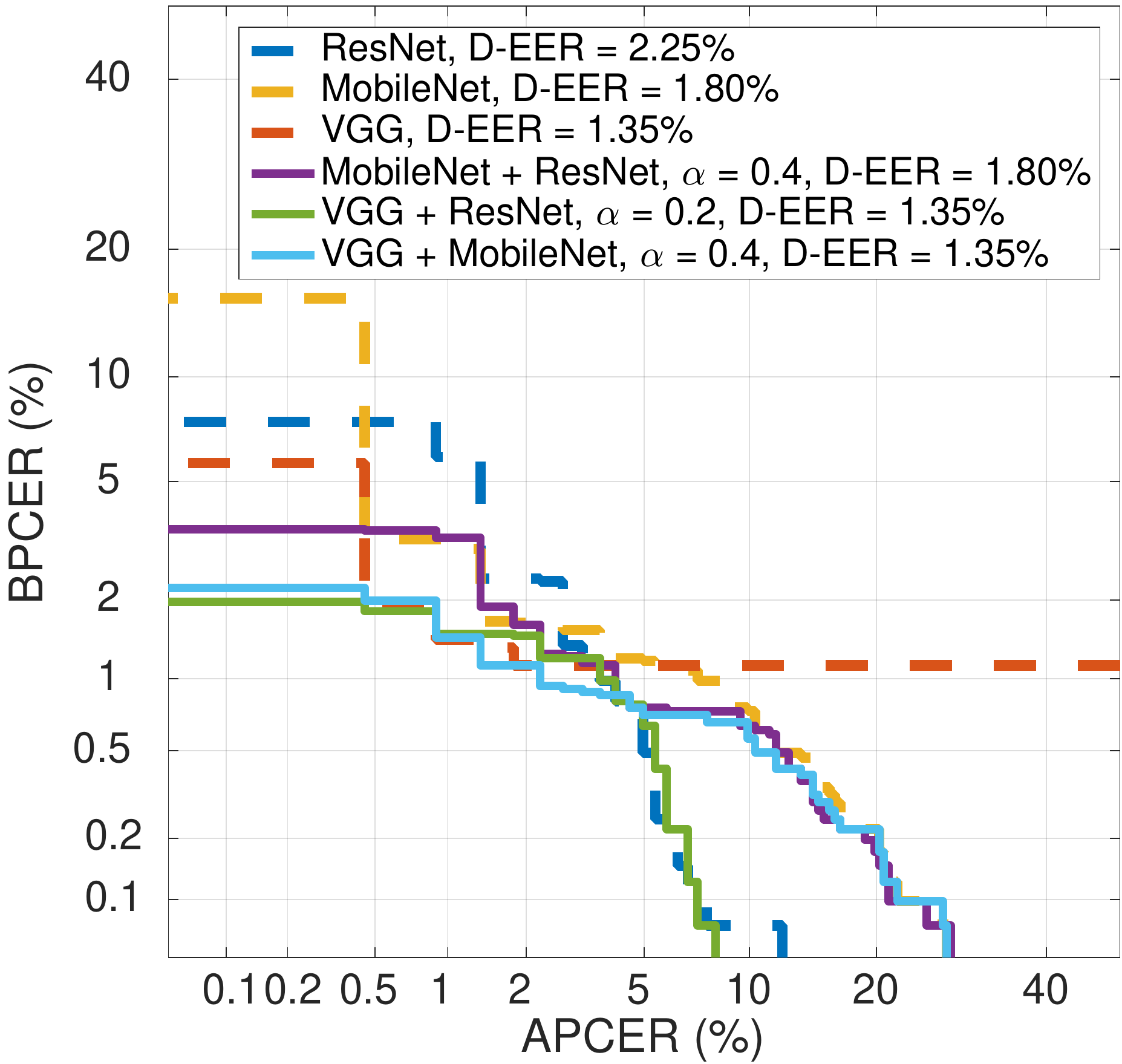}}
\caption{}\label{fig:DET_CNNfusion}
\end{subfigure}
\caption{Performance evaluation of: (a) all the individual systems, (b) the fusion of handcrafted features (SS, Sect.~\ref{sec:pad:svm}) and end-to-end deep learning approaches (MobileNet and VGG19, Sect.~\ref{sec:pad:DL_features}), and (v) the fusion of end-to-end deep learning approaches (ResNet, MobileNet and VGG19, Sect.~\ref{sec:pad:DL_features}).} \label{fig:DET_fusion}\vspace{-0.4cm}
\end{figure*}

For the development of our proposed fingerprint PAD methods, both training and validation datasets are used in order to train the weights of the systems and select the optimal network architectures. For the training dataset, we consider a total of 130 samples for each class (i.e., bona fide and PA), whereas for the validation dataset the number of samples is reduced to 90 per class. It is important to highlight that the same number of samples per class are considered during the development of the systems in order to avoid bias towards one class. 

For the final evaluation, the test dataset comprises the remaining bona fide and PA samples not used during the development of the systems, thereby allowing a fair performance analysis. A total of 4070 and 223 bona fide and PA samples are considered, respectively.

Moreover, it is important to remark that the test dataset includes 5 unkown PAIs, which were not considered during the development stages (i.e., they are not present either in the train or in the validation set). This way, the robustness of the proposed methods to unknown attacks can be evaluated, thereby modelling realistic scenarios. These unknown attacks are underlined in Table~\ref{table:types_PAIs}.

Based on these partitions, three different sets of experiments are carried out:
\begin{enumerate}
\item \textit{Exp 1 - Handcrafted features}: first, the performance of the handcrafted features described in Sect.~\ref{sec:pad:svm} is evaluated. 

\item \textit{Exp 2 - Deep learning features}: then, we evaluate the performance of each deep learning based approach described in Sect.~\ref{sec:pad:DL_features} (i.e., end-to-end and feature-extraction + SVM classification, CNNs trained from scratch and transfer learning), and establish a fair benchmark by following the same experimental protocol.

\item \textit{Exp 3 - Fused system}: in the last set of experiments, the score level fusion (see Sect.~\ref{sec:pad:fusion}) of the aforementioned systems will be evaluated, in order to determine the best performing configuration and assess the complementarity of the individual algorithms.
\end{enumerate}

\section{Experimental Results}
\label{sec:res}


\subsection{Exp 1 - Handcrafted Features}\label{lab:handcrafted_features}

Fig.~\ref{fig:DET_results} shows the DET curves of each of the individual methods proposed in this study. As it may be observed, the spectral signature pixel wise approach has achieved a 12.61\% D-EER. Compared to the results first reported in \cite{Gomez-Barrero-SWIR-SS-PAD-NISK-2018} (APCER = 5.6\% and BPCER = 0\%), there is a clear decrease in the detection performance. This is due to the preliminary character of the first study, over a small database comprising only 60 samples and 12 different PAI species. In this work, the more thorough evaluation unveils the main drawbacks of the approach: it is not possible to get an APCER $\le$ 2\%, and for APCER $\approx$ 5\%, the BPCER is over 20\% (i.e., the system is not convenient any more). 

\subsection{Exp 2 - Deep Learning Features}\label{lab:deepLearning_features}

Deep learning strategies have considerably improved the results achieved using handcrafted features (see Fig.~\ref{fig:DET_results} for a comparison). In general, the features extracted by the neural network models provide a higher discriminative power and generalization to new samples (note that during the development of the systems, all strategies were able to achieve loss values very close to zero for both training and validation datasets). 

For the case of training end-to-end residual CNN models from scratch, the best result obtained is a 2.25\% D-EER. This result outperforms the handcrafted feature approach by relative improvement of 82\%. Furthermore, low APCERs below 1\% can be achieved for BPCERs below 8\%, thereby overcoming the main drawback of the handcrafted features. Similarly, for high convenient systems with BPCERs under 1\%, the APCER ranges between 4 and 15\%. These facts highlight the potential of incorporating residual connections to plain CNNs, being able to easily train neural network models without the necessity of having thousands of labelled images for each class, but only 130 (see Table~\ref{tab:partition_datasets}).   

Very good results have been also obtained for the use of pre-trained CNN models. In particular, the proposed MobileNet- and VGG19-based models have obtained state-of-the-art results with final values of 1.80\% and 1.35\% D-EER, respectively. These results have further improved the results obtained using handcrafted features, achieving an average relative improvement of 86\% and 89\%, respectively.

\begin{figure*}[tb]
\centering
\centerline{\includegraphics[width=\linewidth]{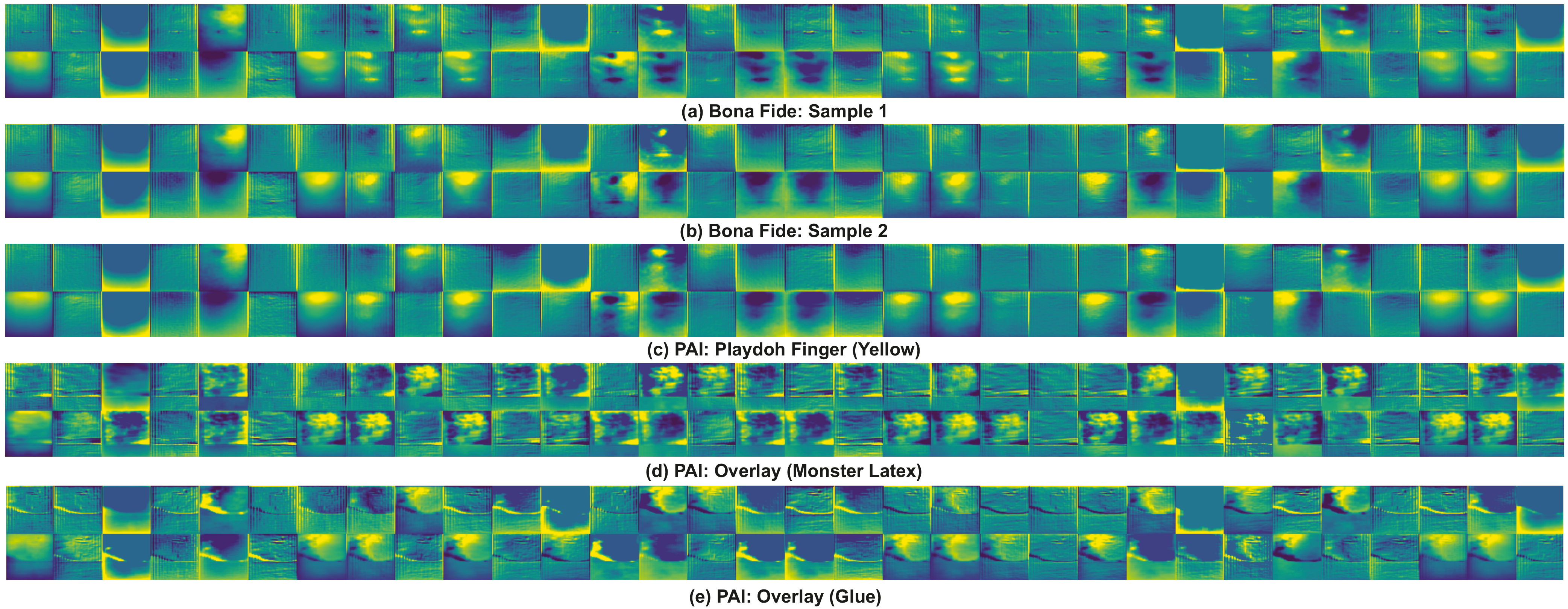}}
\caption{Examples of the features extracted in the first convolutional layer (64 filters) of the VGG19-based model from the samples depicted in Fig.~\ref{fig:input_DNN}.} \label{fig:features_VGG19}\vspace{-0.4cm}
\end{figure*}

In addition, it is important to note that, even though an improvement at the D-EER operating point can be achieved using these end-to-end pre-trained models in combination to transfer learning techniques, with respect to training a network from scratch, this does not hold for all operating points. If we take a closer look at Fig.~\ref{fig:DET_results}, we can observe that for low BPCERs (i.e., high convenience), the best performing approach is the residual CNN trained from scratch. On the contrary, the lowest BPCERs for APCER $\le$ 2\% (i.e., high security) are achieved by the VGG19 pre-trained model. However, it should be noted that the VGG19 based system cannot reach BPCERs under 1\%, which can be done using the pre-trained MobileNet model. Therefore, even if depending on the final application, some CNN approaches might be more suitable than others, the ResNet inspired approach achieves overall the best performance.

For completeness, we also analyse the potential of using CNNs as feature extractors in combination with the SVM classifiers. This way, we can also analyse the improvement achieved using deep learning features compared to the handcrafted features, which were also classified using SVMs. The performance in terms of APCER and BPCER is summarised in Table~\ref{tab:cnn_svm} (note that the SVMs output a single binary decision for the CNN features instead of a score). As it may be observed, the operating points are always contained within the DET curves reported in Fig.~\ref{fig:DET_results}, which means that no further improvement has been achieved using the SVM classification with respect to the last fully-connected sigmoid activation layer of the end-to-end CNNs. Therefore, in the remaining experiments, only the end-to-end CNNs will be considered. On the other hand, the advantages of the learned features with respect to the handcrafted approach are further confirmed.

All these results show the potential of using CNNs in combination with SWIR images for fingerprint PAD purposes, and the robustness of the features extracted. Fig.~\ref{fig:features_VGG19} shows some examples of the features extracted in the first convolutional layer (64 filters) of the VGG19-based model for bona fide and PA samples. In general, very different features are extracted for bona fide and PA samples. This fact can be easily observed when considering overlays based on monster latex and glue, Fig.~\ref{fig:features_VGG19} (d) and (e), respectively. However, the features extracted for the network when considering other materials such as yellow playdoh (Fig.~\ref{fig:features_VGG19} (c)) seem more similar to the bona fide samples (Fig. \ref{fig:features_VGG19} (a) and (b)), indicating the difficulty of the task.

\begin{table}[t]
\small
	\centering
    	\caption{\label{tab:cnn_svm}Performance evaluation of the deep learning feature extractors in combination with the SVM classifiers.}\vspace*{-0.2cm}
	\begin{tabular}{lcc}
		\toprule
		 & BPCER (\%) & APCER (\%)\\
		\midrule
		Residual CNN & 3.37 & 1.35 \\
		MobileNet-Based Model & 5.33 & 0.45 \\
		VGG19-Based Model & 1.89 & 0.90 \\
		\bottomrule
	\end{tabular}\vspace*{-0.4cm}
\end{table}

\subsection{Exp 2 - Deep Learning: Robustness to Unkown Attacks}

Finally, we have also studied the robustness and generalisation capacity of the deep learning methods to new PAIs (a.k.a. unknown attacks). In order to do that, 30 samples acquired from five out of the 35 total PAIs available in the database (see Table~\ref{table:types_PAIs}) were considered only for testing the systems (i.e., none of those PAI samples where included in the training and validation datasets). The reason behind this particular PAI selection is twofold. On the one hand, we chose one PAI species from each row or type on Table~\ref{table:types_PAIs}, to increase the variability also in the unknown attacks. On the other hand, we selected the PAI species with the smallest number of samples available, in order to maximise the number of training samples and hence the detection performance.

In general, very good results have been achieved for all methods. At the D-EER operating point, for the residual CNN and MobileNet-based models only one sample from a yellow playdoh finger has been misclassified, whereas for the case of using the VGG19-based model, all 30 samples stemming from the unknown attacks have been correctly classified. On the other hand, none of the three samples acquired from the yellow playdoh finger were detected by the handcrafted features, which were able to detect the remaining four PAIs. This proves the robustness of the proposed methods to even unknown attacks, which may appear in the future.



\subsection{Exp 3 - Fused Systems}\label{lab:fusion_systems}


In order to further enhance the results achieved by individual methods, and analyse to which degree the systems complement each other, we study in this last set of experiments the fusion of multiple systems at score level. In all cases, the performance has been optimised in terms of the D-EER for values of $\alpha \in [0, 1]$ (see Eq.~\ref{eq:fusion}), where this $\alpha$ weight corresponds to the second system referred to in the legend.

First, the fusion of handcrafted and deep learning features is evaluated in Fig.~\ref{fig:DET_SSfusion}. Only the fusions with the systems based on MobileNet and VGG-19 are depicted, since no improvement was achieved for the fusion of the residual net and the spectral signatures with respect to the individual CNN. As it could be expected given the big performance gap between the spectral signatures based PAD and the deep learning counterparts, the score level fusion yields a minimum improvement with respect to the CNNs only in two cases: $i)$ for either low BPCER $\le$ 0.5\% or low APCER $\le$ 0.5\% for the MobileNet approach (dashed yellow vs solid purple curves), and for $ii)$ BPCER $\le$ 1\% for the VGG19 network (dashed orange vs solid green curves). 

Afterwards, the three CNN based approaches have been fused in a two-by-two basis (the fusion of all three systems showed no further improvement), and the best performing fusions are depicted in Fig.~\ref{fig:DET_CNNfusion}. As it may be observed, no further improvements have been achieved for the operating point around the D-EER. However, for APCER $\le 0.5\%$, the corresponding BPCER values for the fused systems (solid lines) are significantly lower than those of the individual networks (dashed lines): close to 2\% for the fusions with VGG-19 instead of between 5\% and 15\% (i.e., close to a 90\% relative improvement). That yields convenient systems (i.e., low BPCER) even for highly secure (i.e., very low APCER) scenarios. On the other hand, for low BPCER $\le 1\%$, the best APCER ($\le 10\%$) is achieved for either the residual CNN alone (dashed dark blue) or its fusion with the VGG-19 inspired (solid green). In this last case, taking a closer look at the individual PAD scores, we can see that both networks complement each other. Lastly, if we compare Figs.~\ref{fig:DET_SSfusion} and \ref{fig:DET_CNNfusion}, we observe a superior performance in the latter case, thereby further supporting the fact that CNNs can perform better than the baseline handcrafted fusion in this task.

All in all, we can conclude that a remarkable performance can be achieved for fingerprint PAD using SWIR images and the fusion of two CNN models: a residual CNN trained from scratch and a pre-trained VGG-19 CNN. A D-EER as low as 1.36\% can be reached, which is lower to the most similar study in the literature (ACER = 2\% in \cite{chugh-fingerprintPADcnn-TIFS-2018}). Furthermore, other operating points yield a BPCER of 2\% for APCER $\le 0.5\%$, and an APCER $\approx$ 7\% for BPCER = 0.1\%. In addition, the fused system was able to correctly detect all unknown attacks.

\section{Conclusions}
\label{sec:conc}

In this article, we have presented a fingerprint PAD scheme based on $i)$ a new capture device for the acquisition of finger samples in the SWIR spectrum, and $ii)$ state-of-the-art deep learning techniques. An in depth analysis of several networks, either trained from scratch or using transfer learning over pre-trained models, and either as end-to-end solutions or as feature extractors in combination with SVMs for classification, has revealed the soundness of the proposed approach.

Three different CNN architectures have been tested: a residual CNN trained from scratch \cite{DBLP:journals/corr/HeZRS15,DBLP:journals/corr/SzegedyIV16}, and the adaptation of the final layers of the VGG-19 \cite{VGG19_2015} and the MobileNet \cite{DBLP:journals/corr/HowardZCKWWAA17} pre-trained models. In addition, the performance of the proposed DL approaches has been benchmarked against the only handcrafted approach for fingerprint PAD based on SWIR images available in the literature \cite{Gomez-Barrero-SWIR-SS-PAD-NISK-2018}. The performance of all the individual algorithms has been tested over a database comprising more than 4700 samples, stemming from 562 different subjects and 35 different PAI species. Furthermore, several score level fusion schemes have been evaluated. The experimental protocol was designed to simulate a real life scenario: only 260 samples were used for training, and 30 samples acquired from 5 PAI species were excluded from the development stages and utilised only for testing (i.e., unkown attack scenario).

In the aforementioned conditions, the best performance was reached for the fusion of two end-to-end CNNs: the residual CNN trained from scratch and the adapted VGG19 pre-trained model. A D-EER of 1.35\% was obtained. Moreover, this system can be used for different applications. On the one hand, if high user convenience is preferred, an APCER around 7\% can be achieved for a BPCER of 0.1\% (i.e., only 1 in 1000 bona fide samples will be rejected). On the other hand, for highly secure scenarios, a BPCER of 2\% can be achieved for any APCER under 0.5\%. These results clearly outperform those achieved with the handcrafted features, which yielded a D-EER over 12\% and had trouble reaching APCERs under 2\%.

We may thus conclude, that the use of SWIR images in combination with state-of-the-art CNNs offers a reliable and efficient solution to the threat posed by presentation attacks. However, the development of new countermeasures usually brings the corresponding development of new attacks, in this case, new PAI species. To tackle them, we plan to fuse the techniques developed in this work, which analyse the surface of the finger within the SWIR spectrum, with other approaches analysing bona fide properties below the skin \cite{Keilbach-PADlsciTexture-BIOSIG-2018,Kolberg-fingerveinPAD-PLBP-IETBook-2019}.

\section*{Acknowledgements}

\small{This research is based upon work supported in part by the Office of the Director of National Intelligence (ODNI), Intelligence Advanced Research Projects Activity (IARPA) under contract number 2017-17020200005. The views and conclusions contained herein are those of the authors and should not be interpreted as necessarily representing the official policies, either expressed or implied, of ODNI, IARPA, or the U.S. Government. The U.S. Government is authorized to reproduce and distribute reprints for governmental purposes notwithstanding any copyright annotation therein.

This work was supported by the German Federal Ministry of Education and Research (BMBF) as well as by the Hessen State Ministry for Higher Education, Research and the Arts (HMWK) within the Center for Research in Security and Privacy (CRISP, \url{www.crisp-da.de}), and by the projects Cognimetrics (TEC2015-70627-R MINECO/FEDER) and Bio-Guard (Ayudas Fundacion BBVA a Equipos de Investigacion Científica 2017).

This work was carried out during an internship of R. Tolosana at da/sec. R. Tolosana is supported by a FPU Fellowship from Spanish MECD.
}

\bibliographystyle{IEEEtran}
\bibliography{bib/references}

\begin{IEEEbiography}[{\includegraphics[width=1in,height=1.25in,clip,keepaspectratio]{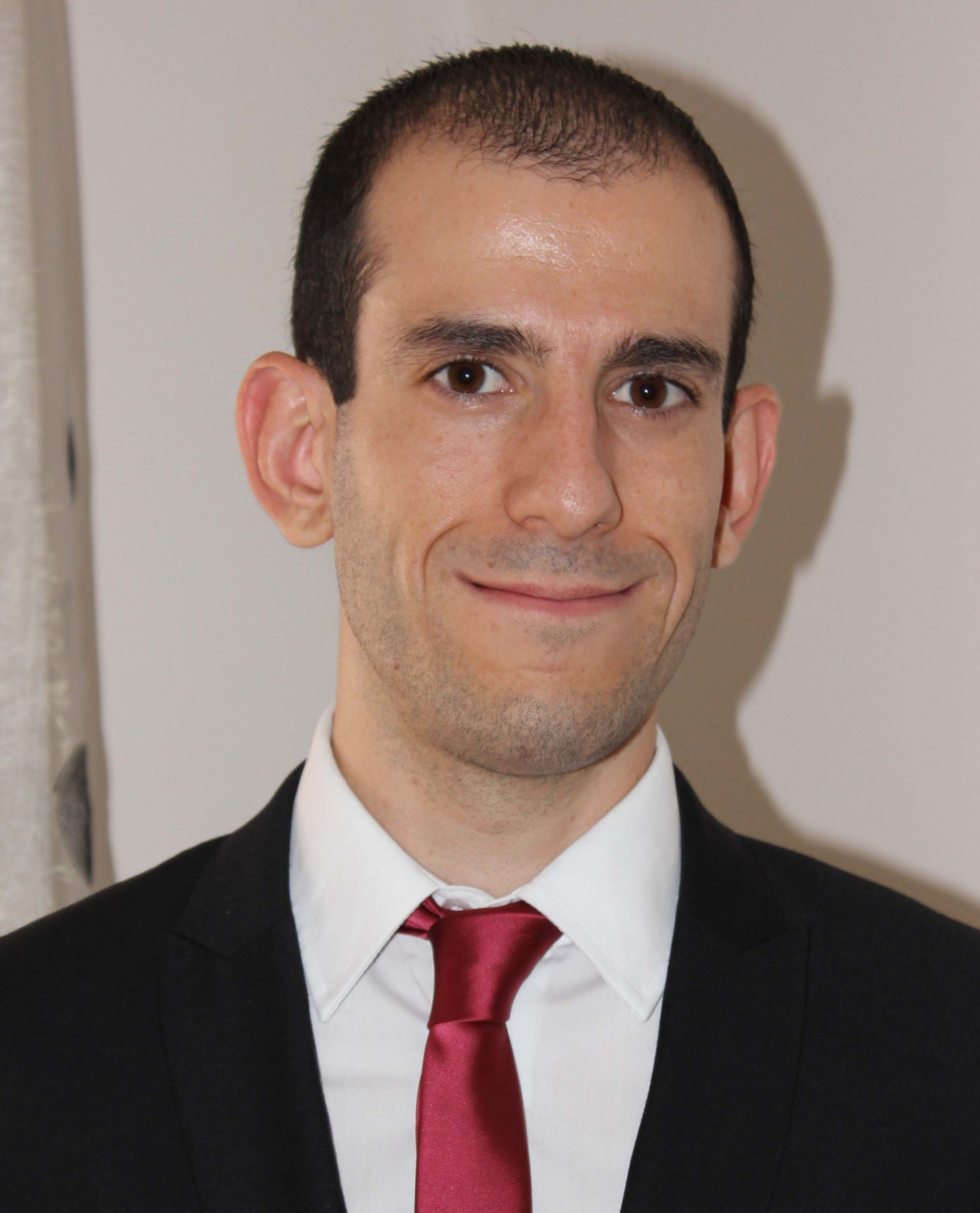}}]%
{Ruben Tolosana}
received the M.Sc. degree in Telecommunication Engineering in 2014 from Universidad Autonoma de Madrid. In April 2014, he joined the Biometrics and Data Pattern Analytics - BiDA Lab at the Universidad Autonoma de Madrid, where he is currently collaborating as an Assistant Researcher pursuing the Ph.D. degree. Since then, Ruben has been granted with several awards such as the FPU research fellowship from Spanish MECD (2015), and the European Biometrics Industry Award (2018). His research interests are mainly focused on signal and image processing, pattern recognition, deep learning, and biometrics, particularly in the areas of handwriting and handwritten signature. He is author of several publications and also collaborates as a reviewer in many different international conferences (e.g. ICDAR, ICB, EUSIPCO, etc) and high-impact journals (e.g. IEEE Transactions of Information Forensics and Security, IEEE Transactions on Cybernetics, ACM Computing Surveys, etc). Finally, he has participated in several National and European projects focused on the deployment of biometric security through the world.
\end{IEEEbiography}

\begin{IEEEbiography}[{\includegraphics[width=1in,height=1.25in,clip,keepaspectratio]{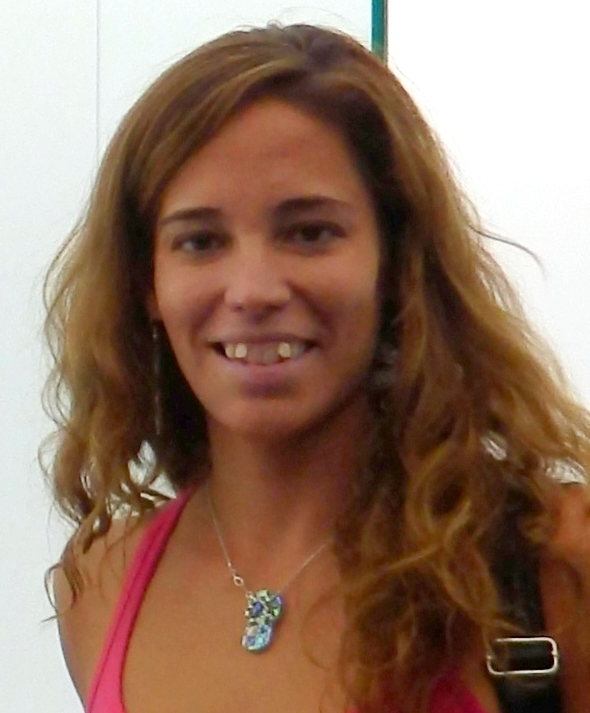}}]%
{Marta Gomez-Barrero}
received her MSc degrees in Computer Science and Mathematics, and her PhD degree in Electrical Engineering, from Universidad Autonoma de Madrid, in 2011 and 2016, respectively. Since 2016 she is a PostDoctoral researcher at the Center for Research in Security and Privacy (CRISP), Germany. Her current research focuses on the development of privacy-enhancing biometric technologies as well as Presentation Attack Detection methods, within the wider fields of pattern recognition and machine learning. She has been actively involved in international projects dealing with vulnerability evaluation of biometric systems, including the EU FP7 projects Tabula Rasa and BEAT, or the BATL project within the US IARPA Odin Program. She is also the recipient of a number of distinctions, including: EAB European Biometric Industry Award 2015, Best Ph.D. Thesis Award by Universidad Autonoma de Madrid 2015/16, Siew-Sngiem Best Paper Award at ICB 2015, Archimedes Award for young researches from Spanish Ministry of Education in 2013 and Best Poster Award at ICB 2013.
\end{IEEEbiography}

\begin{IEEEbiography}[{\includegraphics[width=1in,height=1.25in,clip,keepaspectratio]{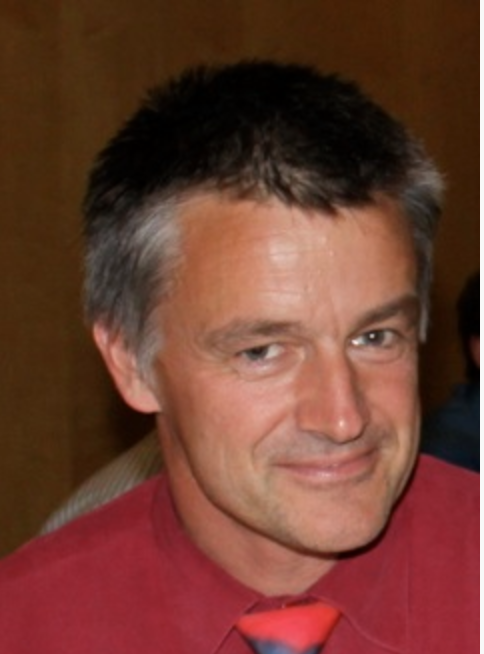}}]%
{Christoph Busch}
received the Diploma degree from the Technical University of Darmstadt (TUD), Darmstadt, Germany, and the Ph.D. degree in computer graphics from TUD, in 1997. He joined the Fraunhofer Institute for Computer Graphics, Darmstadt, in 1997. He is a member of the Faculty of Computer Science and Media Technology with the Norwegian University of Science and Technology, Norway, and holds a joint appointment with the Faculty of Computer Science, Hochschule Darmstadt. Furthermore, he lectures a course on biometric systems with DTU in Copenhagen since 2007. His research includes pattern recognition, multimodal and mobile biometrics, and privacy enhancing technologies for biometric systems. He is Cofounder of the European Association for Biometrics and convener of WG3 in ISO/IEC JTC1 SC37 on Biometrics. He coauthored over 400 technical papers, and has been a speaker at international conferences.
\end{IEEEbiography}

\begin{IEEEbiography}[{\includegraphics[width=1in,height=1.25in,clip,keepaspectratio]{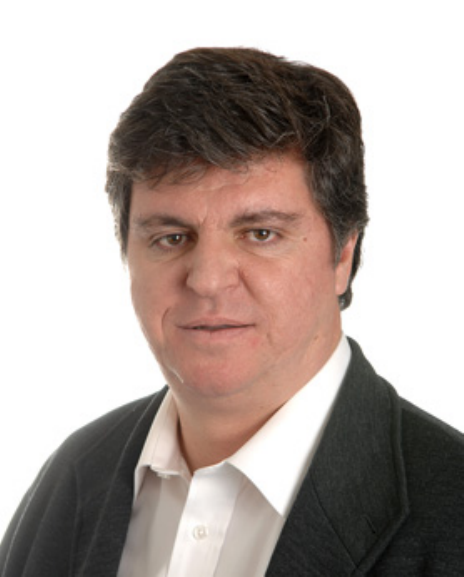}}]%
{Javier Ortega-Garcia}
received the M.Sc. degree in electrical engineering and the Ph.D. degree (cum laude) in electrical engineering from Universidad Politécnica de Madrid, Spain, in 1989 and 1996, respectively. He is currently a Full Professor at the Signal Processing Chair in Universidad Autónoma de Madrid - Spain, where he holds courses on biometric recognition and digital signal processing. He is a founder and Director of the BiDA-Lab, Biometrics and Data Pattern Analytics Group. He has authored over 300 international contributions, including book chapters, refereed journal, and conference papers. His research interests are focused on biometric pattern recognition (online signature verification, speaker recognition, human-device interaction) for security, e-health and user profiling applications. He chaired Odyssey-04, The Speaker Recognition Workshop, ICB-2013, the 6th IAPR International Conference on Biometrics, and ICCST-2017, the 51st IEEE International Carnahan Conference on Security Technology.
\end{IEEEbiography}

\end{document}